\documentclass[twoside]{article}

\usepackage[accepted]{aistats2025}
\usepackage[colorlinks,citecolor=blue,urlcolor=blue,linkcolor=blue,bookmarks=false]{hyperref}
\usepackage{amsfonts,epsfig,graphicx}
\usepackage[linesnumbered,ruled]{algorithm2e}
\usepackage{afterpage}
\usepackage{comment}
\usepackage{amsmath,amssymb,amsthm} 
\usepackage[T1]{fontenc} 
\usepackage{epsf} 
\usepackage{graphics} 
\usepackage{amsfonts,amsmath}
\usepackage[sort]{natbib} 
\usepackage{psfrag,xspace}
\usepackage{color,etoolbox}
\usepackage{wasysym}
\usepackage[export]{adjustbox}
\usepackage{algorithmic}

\def\ind{\perp\!\!\!\perp}

\newcommand{\Pb}{\mathbb{P}}

\usepackage[english]{babel}
\newtheorem{theorem}{Theorem}

\newcommand{\E}{\mathbb{E}}

\DeclareMathOperator*{\argmin}{arg\,min}

\DeclareSymbolFont{bbold}{U}{bbold}{m}{n}
\DeclareSymbolFontAlphabet{\mathbbold}{bbold}

\usepackage{cancel}

\theoremstyle{definition}

\theoremstyle{remark}

\renewcommand\refname{}

\usepackage[utf8]{inputenc} 
\usepackage{hyperref}       
\usepackage{url}            
\usepackage{booktabs}       
\usepackage{amsfonts}       
\usepackage{nicefrac}       
\usepackage{microtype}      
\usepackage{xcolor}         

\begin{document}

%

%

\twocolumn[

\aistatstitle{Accounting for Missing Covariates in Heterogeneous Treatment Estimation}

\aistatsauthor{ Khurram Yamin \And Vibhhu Sharma \And  Ed Kennedy \And Bryan Wilder }

\aistatsaddress{ Carnegie Mellon University } ]

\begin{abstract}
  Many applications of causal inference require using treatment effects estimated on a study population to make decisions in a separate target population. We consider the challenging setting where there are covariates that are observed in the target population that were not seen in the original study. Our goal is to estimate the tightest possible bounds on heterogeneous treatment effects conditioned on such newly observed covariates. We introduce a novel partial identification strategy based on ideas from ecological inference; the main idea is that estimates of conditional treatment effects for the full covariate set must marginalize correctly when restricted to only the covariates observed in both populations. Furthermore, we introduce a bias-corrected estimator for these bounds and prove that it enjoys fast convergence rates and statistical guarantees (e.g., asymptotic normality). Experimental results on both real and synthetic data demonstrate that our framework can produce bounds that are much tighter than would otherwise be possible.
\end{abstract}
\section{Introduction}
Many applications of causal inference require using treatment effects estimated on a study population to make decisions in a separate target population. For example, consider a health system that wishes to deploy a new intervention in their population, using existing study data that was used to estimate heterogeneous treatment effects. The health system will almost certainly have access to features that were not measured in the original study, due to the difference in institutional settings. For example, if the initial study was an RCT, it may have failed to measure practically important covariates \citep{Kahan_Jairath_Dor&eacute;_Morris_2014} such as social determinants of health \citep{Huang2024}. Since the intervention has not previously been used by the health system, no outcome data linked to these new covariates is available. However, treatment decisions would ideally reflect whether the intervention is likely to be beneficial to a patient conditional on \textit{all} information available, not just covariates that happened to be in the original study. This paper studies the question: how precisely can we identify treatment effects conditional on such new covariates? If precise estimates are available, the decision maker can proceed confidently with deployment. Conversely, if considerable uncertainty remains about an important subgroup, a decision maker may exercise more caution or invest more resources in monitoring or additional data collection. 

Formally, we aim to derive bounds on conditional average treatment effects (CATEs) when novel covariates are observed in the target population. We refer to the CATE conditional on both the common and new covariates as the \textit{fully conditional} CATE and the CATE conditional on only the common covariates (which is what can be estimated from the original study) as the \textit{restricted} CATE. Intuitively, what makes informative bounds possible is that the fully conditional CATE must be consistent with restricted CATE when marginalized to only the common covariates. This idea is reminiscent of the ecological inference literature, which focuses on inferring a joint distribution from its marginals. Ecological inference has long been used in the quantitative social sciences, e.g.\ for election analysis \citep{Glynn_Wakefield_2010,King_Rosen_Tanner_Tanner}. However, almost no previous work uses ideas from ecological inference in causal settings. We provide a partial identification strategy new to causal inference by connecting ideas from the ecological inference to causality. The resulting bounds on treatment effects use the joint distribution of the common and new covariates to link the fully conditional and restricted CATEs. In the worst case, if these sets of covariates are entirely independent, we cannot hope for very strong bounds on treatment effects. However, if they are strongly associated (as in many practical settings), the fully conditional CATE must be anchored more strongly to particular values of the restricted CATE.



We make the following contributions. First, we provide formally provably bounds on conditional treatment effects by leveraging ideas from ecological inference. These bounds contain nuisance functions that must be estimated. Our second contribution is a bias-corrected estimator that exhibits favorable statistical properties such as allowing for the use of non-parametric and/or slow converging machine learning models to estimate these nuisance functions without sacrificing fast $O_{\mathbb{P}}(\frac{1}{\sqrt{n}})$ rates convergence rates. We also prove that our estimator is asymptotically normal, facilitating the construction of confidence intervals. Finally, we demonstrate these properties empirically through the use of simulation and application to data from a real RCT.

\textbf{Additional related work: }
There is a great deal of work that focuses on combining experimental and observational data to estimate treatment effects. This paper does \textit{not} focus on the distinction between experimental and observational data: our setup is agnostic as to whether the study population is experimental or observational as long as it satisfies standard identification assumptions. Our focus is on incorporating \textit{covariates} that are newly observed in the target population and not present in the study. One major line of previous work attempts to use outcome data from both a RCT and observational study to jointly estimate treatment effects (on common covariates). Often, this involves fitting a model for the confounding bias present in the observational study \citep{Kallus_Puli_Shalit_2018,Yang,Wu_Yang_2022}. By contrast, we focus on the setting where no outcome data is available in our target group and new covariates are present. A second line of work focuses on the case where the observational study has no outcomes
by correcting for shift in the covariate distribution \citep{Lesko_Buchanan_Westreich_Edwards_Hudgens_Cole_2017, Lee_Yang_Wang_2022}. These methods deal only with covariates found in common, and focus on average effects (as opposed to our focus on conditional effects).


Recently, a few papers have tried to quantify the uncertainty from covariates missing in the RCT. \cite{50144} focus on quantifying how an estimate of the ATE might be biased when a covariate is missing in one or both populations. In contrast, our focus is on bounding conditional treatment effects. They also require distributional assumptions (e.g., Gaussianity) on the missing covariate, as opposed to our nonparametric approach. Similarly, \cite{Nguyen_Ebnesajjad_Cole_Stuart_2016} propose a sensitivity analysis framework for estimation of the ATE in a target population when a treatment effect mediator is unobserved in the target population (as opposed to our focus on covariates unobserved in the study population).  \cite{Andrews_Oster_2019} propose a framework to assess the external validity of RCTs when a missing covariate induces selection bias into the trial. To our knowledge, no previous identifies non-parametric bounds on conditional treatment effects for generalizing treatment effects to target populations with unaccounted for covariates.

\section{Problem Setup}
We examine the situation where we are attempting to transport estimates of heterogeneous treatment effects between two populations. In the first population, which we refer to as the \textit{study} population, we observe covariates $V$, treatment assignments $A \in \{0, 1\}$, and outcomes $Y$. Each individual has potential outcomes $Y^1$ and $Y^0$ that would be realized if they were (respectively, were not) treated. We assume that $Y^1, Y^0 \in [a,b]$ with probability 1 for some constants $a$ and $b$, i.e., the outcomes are bounded between known values. We observe $Y^A$ corresponding to the treatment assignment. In this population, we impose standard identifying assumptions (most prominently, no unobserved confounding) that allow estimation of the conditional average treatment effect (CATE) $\E[Y^0 - Y^1|V]$. This study population could represent a randomized experiment or an unconfounded observational setting. This assumption is formalized as $A \ind Y^a \mid V$ given the context of the study population. 

In the second population, which we refer to as the \textit{target} population, we do not observe the treatment or outcome variables. Instead, we observe just the covariates, which consist of both $V$ and a new set of  covariates $W$ which were not observed in the study.  We assume that $W$ consists only of discrete covariates (although $V$ may be either discrete or continuous). This holds naturally in many settings of practical interest (e.g., social determinants of health are very often discrete variables \citep{lwwUnderstandingData}), or can otherwise be ensured via discretization (e.g., many clinical risk scales are already discretized into a fixed set of levels \citep{ustun2019learning}). This assumption is technically required so the probability of a specific realization of $W$ is well-defined. 
We use an indicator variable $E$ to indicate whether the subject is in the study population ($E = 1$) or the target population ($E = 0$). Our observed data consists of samples
$$ Z=\Big\{ V, E, W(1-E), E(A, Y) \Big\} = \begin{cases} V,W & : E=0 \\ V,A,Y & : E=1  \end{cases}  $$
Our goal is to estimate $\E[Y^1 - Y^0|V, W]$, or the CATE conditioned on both $V$ and $W$. We refer to this quantity as the \textit{fully conditional} CATE. 
 We are going to use the standard assumptions of consistency (Y=$Y^a$ whenever A=a), positivity ($\Pb(A=a|V=v,W=w)>0 $ for all combinations of v,w). We also assume that
(1) $\Pb(W=w \mid V=v, E=1) = \Pb(W=w \mid V=v, E=0) $, and
(2) $\E(Y^1-Y^0 \mid V,W, E=1) = \E(Y^1-Y^0 \mid V,W, E=0)$. These  two assumptions could be restated as (1) the covariates missing in the study population have the same conditional distribution as those in the target population, and (2) the fully conditional CATE ($\E[Y^1-Y^0 \mid V,W]$)  is the same in the study and target population.

\section{Methodology}
\subsection{Partial identification bounds}
Our goal is to provide as much information as possible about the fully conditional CATE (i.e., conditional on both $V$ and $W$) when outcome data is linked only to $V$. Clearly, in this setting it is not possible to exactly identify the fully conditional CATE. However, we can use ideas from ecological inference to \textit{partially} identify it. Specifically, if we average the fully conditional CATE over values of $W$, we must obtain the CATE conditioned only on $V$, which \textit{is} identified. Formally, a long line of work in ecological inference \citep{Jiang_King_Schmaltz_Tanner_2020,Plescia_De_Sio_2017, RePEc:nbr:nberwo:22643} uses marginal consistency conditions of the following form, for a single outcome $Y$:
\begin{align}
    E(Y \mid v) = \sum_w E(Y  \mid v, w) p(w \mid v). \label{eq:ecological}
\end{align}
For the sake of brevity, we use capital letters (eg $V,W$) to represent random variables and lowercase versions (eg $v,w$) to represent specific values of these letters (such that $V=v, W=w$). If $p(w \mid v)$ is known or can be estimated, we can then rearrange this expression to obtain bounds on $E(Y  \mid v, w)$. We apply a similar strategy in the context of the CATE (using also the causal assumptions that link it to the observable data) and obtain the following provable bounds on the fully conditional CATE:
\begin{theorem} \label{theorem:identification}
Assuming Y is a real-valued outcome bounded in [a,b]
\begin{align*}
    \gamma_\ell(v,w) \leq \E(Y^1-Y^0 \mid v,w,E=1) \leq \gamma_u(v,w)
\end{align*}
such that
\begin{align*}
\gamma_\ell(v,w) &= \max\left\{ \frac{ \mu_1(v) - \mu_0(v)  - (b-a) (1-\nu)  }{ \nu }  , a-b \right\} \\
\gamma_u(v,w) &= \min\left\{ \frac{ \mu_1(v) - \mu_0(v)  - (a-b) (1-\nu)  }{ \nu }  , b-a \right\}
\end{align*}
where 
$ \mu_a(v) = \E(Y \mid V=v, A=a, E=1),\\ 
\nu(v,w) = \Pb(W=w \mid V=v, E=0)  $
\end{theorem}
This bound utilizes the worst-case scenario that if $Y \in [a,b]$, then $Y^1-Y^0 \in [a-b, b-a]$.

Note that these bounds (our estimand of interest) are a function of $V$ and $W$. We introduce a framework which estimates the projection of these functions onto a parameterized class chosen by the analyst. Different use cases may call for different parameterizations. E.g., an analyst may prefer a linear model for simplicity and easy of interpretation, or use a more intricate nonlinear model such as a neural network if they wish for greater flexibility to capture the underlying function. Regardless of the choice, we assume that the model is parameterized by some $\beta$ of fixed dimension with respect to $n$. Note we do not assume that the model $m$ is correctly specified; i.e.\, we estimate the projection of $\gamma$ onto a model class instead of setting $\gamma$ equal to some model.


For simplicity of notation, we will use $X$ interchangeably with the joint tuple $(V,W)$. We attempt to find the best approximation of $\gamma_\ell(x)$ and $\gamma_u(x)$  in the form of a model
$m$. We use $\gamma$ to represent the bound of interest, which could be either $\gamma_\ell(x)$ or $\gamma_u(x)$ as desired. We seek model parameters $\beta$ which minimize the mean squared error between $m$ and $\gamma$: 
\begin{align*}
\beta &= \argmin_{\theta \in \mathbb{R}^p} \mathbb{E} \left[ h(X) \{\gamma(X) - m(X; \theta)\}^2 \right] 
\end{align*}
This formulation is weighted by an analyst-chosen function $h$ which allows a degree of choice on what observations to place emphasis on (for example we could choose to emphasize certain covariate regions of interest).
Differentiating then gives us the moment condition $M(\beta)$ 
\begin{align*}
0 &= \mathbb{E} \left[ \frac{\partial m(X; \beta)}{\partial \beta} h(X) \{\gamma(X) - m(X; \beta)\} \right] = M(\beta) 
\end{align*}
Our goal becomes to find the $\hat{\beta}$ such that 
$M(\hat{\beta})= 0$. To keep notation concise, define 
\begin{align*}
   \tau_\ell(x) = \frac{ \mu_1(v) - \mu_0(v)  - (b-a) (1-\nu)  }{ \nu } 
\end{align*}
so that $\gamma_\ell(x) = \max\{\tau_\ell(x),a-b\}$, and similarly for $\tau_u(x)$. Letting $g(X) =\frac{ \partial m(X; \beta)}{\partial \beta} h(X)$, this moment condition is equivalent to 
\begin{align}
    M_\ell(\beta_\ell)  = 
    E[g(X)\{(\tau_\ell(X)+b-a)* \notag \\  \mathbf{1}(\tau_\ell(X)+b-a \geq 0) 
    +a-b -m(X; \beta) \}  ] = 0 \label{eq:moment-condition-lower} \\
    M_u(\beta_u)  = 
    E[g(X)\{(\tau_u(X)+a-b)* \notag \\
    \mathbf{1}(\tau_u(X)+a-b \leq 0)+b-a -m(X; \beta) \}  ] = 0 
\end{align}

\subsection{Estimation} \label{section:estimation}
If $\tau$ were known, we could simply solve the resulting least-squares problem. However, $\tau$ in fact depends on a number of nuisance functions that are not known and must be estimated from the data: $\hat{\mu}_1(V), \hat{\mu}_0(V)$, and $\hat{\Pb}(W=w \mid V=v, E=0)$. A naive plug-in strategy would be to estimate each of the nuisance functions and then plug the estimates into the moment condition. For example, for a linear model class $m(v,w,\beta)= \beta_1^Tv + \beta_2^Tw$, we would obtain a plug-in estimator of:
\begin{align} \label{eq:plugger}
    \hat{\beta_\ell}  &= 
    \mathbb{P}_n\{ XX^T\}^{-1} \mathbb{P}_n\{ X\{(\hat{\tau}_\ell(X)+b-a)* \\& \mathbf{1}(\hat{\tau}_\ell(X)+b-a \geq 0)+a-b \} \label{eq:plugger2}\\
    \hat{\beta}_u &= 
    \mathbb{P}_n\{ XX^T\}^{-1} \mathbb{P}_n\{ X\{(\hat{\tau}_u(X)+a-b)*\\& \mathbf{1}(\hat{\tau}_u(X)+a-b \leq 0)+b-a\}
\end{align}
where $\frac{\partial m(v, w; \beta)}{\partial \beta} =  \begin{bmatrix}
V & W
\end{bmatrix}^T = X$.

However, the quality of the resulting solution will depend sharply on how well the nuisances are estimated. In general we will not obtain consistent estimates for even the projection onto the parametric model class unless the nuisances are estimated consistently. As the nuisances are unlikely to lie exactly in any specific parametric class, consistent estimation will require the use of nonparametric methods that converge only slowly (slower than $O(n^{-\frac{1}{2}})$). Conversely, if the true values of the nuisance functions were known, $\beta$ could be estimated at the (faster) root-$n$ parametric rate \citep{Vaart_2000}. We draw on techniques from the semiparametric statistics/double ML literature to propose a Bias-Corrected estimator that attains the parametric rate for $\beta$ even when the nuisance functions are  estimated at slower nonparametric rates.

\subsection{Bias-Corrected Estimator}
In this section, we will focus on the derivation of a Bias-Corrected estimator for $\gamma_\ell(x)$; $\gamma_u(x)$ follows a similar form, shown in the Appendix. Full proofs of all claims can be found in the Appendix. The starting point is to derive an \textit{influence function} for our estimand of interest. Intuitively, influence functions approximate how errors in nuisance function estimation impact the quantity of interest, in this case $\gamma$. A common strategy in semiparametric statistics is to use the influence function for the target quantity to provide a first-order correction for the bias introduced by nuisance estimation. This dampens the sensitivity of the estimator to errors in the nuisances and will allow us to derive fast convergence rates for $\beta$ even when the nuisances converge more slowly.


Unfortunately, influence functions typically only exist for quantities that are pathwise differentiable. The expression for $\gamma_\ell$ contains a non-differentiable max, which shows up as an indicator function in the moment $M_\ell(\beta_\ell)$ (\ref{eq:moment-condition-lower}). That is, the moment condition is discontinuous on the margin $\tau(x)+b-a = 0$ to $\tau_\ell(x)+b-a < 0$. We employ a margin condition strategy used by \citet{kennedy2019sharp}, who analyzed a nondifferentiable instrumental variable model, and others \citep{pmlr-v151-kpotufe22a, vigogna2022multiclass}. Specifically, we can hope for fast estimation rates when the classification problem of estimating the indicator for a given $X$ is not too hard, in the sense that not too much probability mass is concentrated near the boundary. Specifically, we assume that  \\
\text{\bf Assumption (Margin Condition):}
\begin{align}
    \label{result:margin}   P(|\tau_\ell(x)+b-a|\leq t) \leq Ct^\alpha
\end{align} 
for some constant C and some $\alpha \geq 0$. Similar assumptions are often imposed in the context of classification problems \citep{Audibert_2007}. We have $\alpha=0$ with no further assumptions, and $\alpha=1$ holds if $\tau(x)$ has a bounded density. A bounded density as a relatively weak assumption, so $\alpha \geq 1$ is likely to hold in many cases of interest.

Under this margin condition, we employ a two-part strategy. First, we derive an influence function for the moment condition under the assumption that the true value of the indicator function is known, rendering the expression differentiable in the estimated nuisances. Second, our final estimator replaces the indicator function with a plug-in estimate; the margin condition entails that this step introduces relatively small bias compared to if the true indicator were known. Specifically, we prove that 
\begin{align*}
    \mathbb{E}[\tau_\ell\mathbf{1}(\hat{\tau}_\ell(X)+b-a) -\tau_\ell \mathbf{1}(\tau_\ell(X)+b-a) ] \\
      \leq C ||\hat{\tau}_\ell(X)-\tau_\ell(X)||_{\infty}^{1+\alpha} 
\end{align*}
When $\alpha \geq 1$, this term now depends only on squared errors in the estimation of $\tau$ and will be negligible asymptotically so long as $\hat{\tau}_\ell$ converges at a $o(n^{-\frac{1}{4}})$ rate. The proof for this is contained in the appendix. For simplicity of notation, we call the indicator $f(X)$ such that $f(X) =\mathbf{1} (\tau(x)+b-a \geq 0)$. 

Next, we turn to deriving the influence function for $\gamma$ under the assumption that the indicator $f$ is known (where our eventual estimator will use the analysis above to justify replacing $f$ with its plugin estimate). 
Given this strategy, we derive $\varphi(X,\beta,\eta)$ (full  derivation in Appendix) to obtain:
\begin{theorem}  \label{theorem: Influence}
$\varphi(X,\beta,\eta)$ is given by:
\begin{align*}
 &\sum_w  \begin{bmatrix}
V \\
w 
\end{bmatrix}f(V,w)  \{\frac{EA}{\pi_1(V)}(Y-\mu_1(V))
\\ & - \frac{E(1-A)}{\pi_0(V)}(Y-\mu_0(V))\}
    \\
&+ \frac{(b-a)(1-E)}{\rho_0(V)} \{ 
X f(X)  - \sum_w  (f* \nu)(V,w)
\} \\
& - \frac{1-E}{\rho_0(V)} \{   
X f(X) \tau_\ell(X) - \sum_w \begin{bmatrix}
V \\
w 
\end{bmatrix} (f* \tau_\ell*\nu)(V,w) \} \\
&+ X \{\tau_\ell(X)f(X)+(b-a)f (X)+a-b - m(X,
\beta) \} 
\end{align*}
where $\pi_a(V)  = p(A=a,E=1|V)$, \\
$\rho_0(V)= p(E=0|V)$
\end{theorem}
From a inspection of the influence function, it can be seen how the bias correction emerges. There is a strong similarity in structure of the plug-in formulation (Equations \ref{eq:plugger},\ref{eq:plugger2}) to the last line of $\varphi$. The first 4 lines of $\varphi$ serve to correct for misspecification in the plug-in estimator. For example, the first two lines of $\varphi$  corrects for misspeficiation in the $\mu_1(V) - \mu_0(V)$ portion of $\tau_\ell$ if the propensity scores $\pi_a(V)$ and $\rho_0(V)$ are correctly specified. $\pi$ and $\rho$ give the probabilities of being treated and/or being in the study population, respectively. These appear as additional nuisances in the  influence function; including these additional nuisances will give our estimator greater robustness to estimation errors.

From this influence function, we can construct our bias-corrected estimator. The key idea is to find $\hat{\beta}$ that solves an estimating equation implied by the influence function:
    \begin{align*}
      \hat{\beta} \text{ such that }
      P_n\{\varphi(X,\hat{\beta},\hat{\eta})\}=0
    \end{align*}
Formally, in order to construct the estimator, we employ a sample splitting procedure detailed in Algorithm \ref{algo: alg}. This follows the strategy, common in semiparametric inference \cite{kennedy2021semiparametric}, of splitting the dataset into two independent halves. The first is used to estimate the nuisance functions (including the indicator $f$). Then, we fix the nuisances and construct $\varphi(X,\hat{\beta},\hat{\eta})$ for the points in the second half, which are used to estimate $\beta$ via the above moment condition. The computational approach to solving the moment condition will depend on the model family chosen. For linear models $m(v,w,\beta)= \beta_1^Tv + \beta_2^Tw$, we give a closed-form solution in the appendix that can be computed as a standard OLS problem. For more general differentiable model classes, one strategy would be to minimize $P_n\{\varphi(X,\hat{\beta},\hat{\eta})^2\}$ using gradient-based methods.

We now turn to analyzing the convergence properties of the bias-corrected estimator, with the goal of showing that $\hat{\beta} \to \beta$ at a fast rate even when the nuisances are estimated slowly. We start by examining the bias $R_n$ of our influence function. Let $\eta_0$ be the true values of the nuisance functions while $\hat{\eta}$ is our estimate. The bias $R_n$ quantifies the difference between the expected influence function at $\eta_0$ and $\hat{\eta}$ and plays a key role in controlling the convergence rate of $\hat{\beta}$. Formally, we decompose the bias as:
\begin{theorem}  \label{theorem: Bias}
    Let  $R_n = \mathbb{P}\{\varphi(X; \beta, \hat{\eta}) - \varphi(X; \beta, \eta_0)\}$. Assuming that all nuisance functions and their estimates are bounded below by a constant larger than 0 and that all probabilities are bounded above by 1,\begin{align*} 
R_n &\mathrel{\substack{\le \\ \sim}} \| {\hat{\rho}}- \rho    \|_2
\|   \hat{\nu}-\nu   \|_2 +
\|  \hat{\nu}- \nu
    \|_2^2 \\
    &+ \| \hat{\pi}_1 -\pi_1\|_2 \|\hat{\mu}_1 -\mu_1 \|_2  
     +  \| \hat{\pi}_0-\pi_0 \|_2 \| \hat{\mu}_0 - \mu_0 \|_2 \\
     &+ \|  \mu_1-\hat{\mu}_1 \|_2 \|  \hat{\nu}-\nu \|_2 
    +
  \| \hat{\mu}_0- \mu_0 \|_2 \| \hat{\nu}-\nu\|_2 
    \end{align*}
\end{theorem}
Roughly, our estimated $\hat{\beta}$ will converge quickly if $R_n = o_{\mathbb{P}}(\frac{1}{\sqrt{n}})$ (a statement formalized below). For this to occur, all of the products above must be $o_{\mathbb{P}}(\frac{1}{\sqrt{n}})$. It is a sufficient but not a necessary condition that $ \hat{\rho}  ,  \hat{\nu} ,\hat{\pi}$, and $\hat{\mu}$
converge to their true functions at $o_{\mathbb{P}}(n^{-\frac{1}{4}})$ rates for  $R_n$ to be $o_{\mathbb{P}}(\frac{1}{\sqrt{n}})$. Note that $n^{-\frac{1}{4}}$ is substantially slower than the parametric root-$n$ rate, and is satisfied by even many nonparametric methods. It becomes clear from the bias structure that $\hat{\nu}= \hat{P}(W=w|V=v,E=0)$ would be need correctly specified for $\hat{\beta}$ to be a consistent estimator of $\beta$. This is due to the nature of the ecological inference setting where $\hat{\nu}$ is the only thing linking the target population to the study. Thus, our estimator is not doubly robust in the sense that no other nuisance can compensate for errors in $\hat{\nu}$. However, the error is still second-order as it involves only squared errors for $\nu$. Additionally, there is a mixed-bias property with respect to all of the other nuisances, where each nuisance can converge at a slower rate individually if others converge faster (e.g., $\mu$ can converge more slowly if $\pi$ and $\nu$ converge faster). Therefore, our estimator exhibits robustness to misspecification in all nuisances except $\nu$. Putting these pieces together, we obtain a convergence guarantee for $\hat{\beta}$:

\begin{theorem}  \label{theorem: Convergence}
    If $R_n = O_{\mathbb{P}}(\frac{1}{\sqrt{n}})$ and assuming the margin condition (\ref{result:margin}) holds for $\alpha \geq 1$:
    \begin{align*}
    ||\hat{\beta}-\beta||_2 = O_{\mathbb{P}}\left(\frac{1}{\sqrt{n}}\right) \hspace{1cm}
    \end{align*}
    and
    \begin{align*}
    \sqrt{n} (\hat{\beta} - \beta) \xrightarrow{\text{D}} N\big(0,  M^{-1}VM^{-1}) \big)
\end{align*}
where $M=\frac{\partial \mathbb{E}(\varphi(X; \beta, \eta_0))}{\partial \beta^T}$ and V is the variance of $\varphi(X; \beta, \eta_0)$
\end{theorem}

Theorem 4 follows from Theorem \ref{theorem: Bias} combined with a standard analysis of $M$-estimators under misspecification (c.f.\ Theorem 5.2.1 of \cite{Vaart_2000} and Lemma 3 of \citet{kennedy2021semiparametric}).  Asymptotic normality is valuable as it allows for the construction of confidence intervals, e.g.\ with the usual sandwich estimator (or potentially easier in practice, the bootstrap).
\begin{algorithm} \label{algo: alg}
\caption{Algorithm for Constructing Bias-Corrected Estimator}

 Given input samples $\mathcal{D}$, split uniformly at random into $\mathcal{D}_1$ and $\mathcal{D}_2$ 
 \\
 Use $\mathcal{D}_1$ to estimate the nuisance functions: $\hat{\tau}_\ell(X)$, $\hat{\mu}_1(X)$, $\hat{\mu}_0(X)$, $\hat{\pi}_1(X)$, $\hat{\pi}_0(X)$, $\hat{\rho}_0(X)$, $\hat{\nu}(X)$\\

Use the estimated nuisances estimates to construct the estimated indicator: $\hat{f}(X) = \mathbf{1}(\hat{\tau(x)} + b - a \geq 0)$ using the first part of the data. Let $\hat{\eta}$ be the set of all nuisance estimates, including now $\hat{f}(X)$. \\

Find $\hat{\beta}$ that solves the following estimating equation on $\mathcal{D}_2$:
    \[
    \frac{1}{|\mathcal{D}_2|}\sum_{X \in \mathcal{D}_2} \varphi(X, \hat{\beta}, \hat{\eta}) = 0
    \]
    where $\varphi(X, \hat{\beta}, \hat{\eta})$ is constructed using $\hat{\eta}$ following the expression in Theorem \ref{theorem: Influence}.
    
 Output $\hat{\beta}$
\end{algorithm}

\subsection{Bounds in a sensitivity model} \label{section:sensitivity}

In some cases, we may be willing to impose additional assumptions limiting the deviation between the restricted and fully conditional CATEs. This may be justified based either on domain knowledge, or taken in the spirit of a sensitivity analysis where the analyst varies a parameter controlling the strength of such assumptions to see how much variation across levels of $W$ their conclusions are robust to. In this section, we propose such a sensitivity analysis model to formalize the case where $W$ is believed to have a limited impact on treatment effects, after conditioning on $V$. Specifically, our sensitivity model imposes the assumption that 
\begin{align}
    \big| \E(Y^1 - Y^0 \mid V&=v,W=w, E=1) \notag \\
     - \E(Y^1 - Y^0 \mid V=v, E=1) \big| \leq \delta 
\label{eq:sensitivity}
\end{align}
or that the fully conditional effects cannot differ from partly conditional effects by more than $\delta$. $\delta$ here is a user-chosen parameter, that may be varied to test the robustness of estimates to an increasingly strong effect of $W$. At $\delta = 0$, the fully conditional CATE is equal to the restricted CATE, and at $\delta = b - a$, we recover our previous bounds (that use only the boundedness of the outcome to $[a,b]$). At any intermediate level of $\delta$, we obtain partial identification bounds that are of a similar form to those in Theorem \ref{theorem:identification} (but stronger), simply replacing the terms $(a-b)$ or $(b-a)$ by $ \E(Y^1-Y^0 \mid V=v, E=1) \pm \delta$. We estimate these bounds using the same strategy as for the original model, plugging in the identified quantity $\E(Y^1-Y^0 \mid V=v, E=1) = \mu_1(v) - \mu_0(v)$ using data from the studyal population.


\section{Experiments}
\textbf{Setup: }
We start our experimentation with experiments on simulated data so that the ground-truth CATEs are known, and afterwards give an application on data from a real RCT. Full  details of the simulation are given in Appendix. The process samples 10,000 observations of 3 continuous covariates for $V$ and 3 discrete (binary) covariates for $W$. $W$ is simulated as $\Pr(W_i = 1) = \text{logit}(\alpha^T_i V)$ for a coefficient vector $\alpha_i$, where the choice of $\alpha$ allows us to control the degree of dependence between $V$ and $W$. $E$ and $A$ are generated similarly as functions of $V$, producing covariate shift between the populations and nonuniform assignment to treatment across levels of $V$ within the study population. Finally, the CATE is a specified function of both $V$ and $W$ and we sample observed outcomes from the study population matching the CATE. 

\textbf{Baselines: }To our knowledge, ours is the first paper to present an algorithm to identify and estimate CATE bounds for covariates unobserved in the original study. We compare to three baselines. First, we compare the performance of our bias-correct estimator to the plug-in estimator discussed in Section \ref{section:estimation}, which uses the same nuisance function estimates as our model but estimates $\beta$ directly by plugging these estimates into the moment condition, without the influence function-based correction. Second, we compare the informativeness of our bounds to those which use all of our assumptions \textit{except} the key ecological inference component in Equation \ref{eq:ecological}, to test whether using this information results in tighter identification (more details on this below). Third,  we compare the frequency with which our bounds cover the true treatment effects for specific subgroups of $(V,W)$ with the coverage of confidence intervals for the restricted CATE (which is measured on $V$ and not $W$). To do this, we implement a standard doubly-robust estimator for the restricted CATE \citep{kennedy2023optimaldoublyrobustestimation}, with the same model families as our own estimator.

\textbf{Benchmarking: }In order to set the sensitivity parameter $\delta$, we can use benchmarking methods similar to other sensitivity models in causal inference. \citep{byun2024auditingfairnessunobservedconfounding,Hosman2010,mcclean2024calibratedsensitivitymodels}. For a specific instance, the analyst holds out n variables from $V$  to form a ``simulated" set of additional covariates $W'$ mimicking $W$ where $V'$ represents the original $V$ excluding $W'$. They then use the difference $|\E[Y^1-Y^0|V'] - \E[Y^1-Y^0|V',W']|$  as a proxy for  $|\E[Y^1-Y^0|V] - \E[Y^1-Y^0|V,W]|$ in order to set $\hat{\delta}$. This procedure is performed on all subsets of n variables in $V$ and results are averaged. As in sensitivity analyses for unmeasured confounding, this provides an interpretation (based on domain knowledge) that the analysis is robust to variables ``at least as important" as $W'$.

\begin{figure} 
    \centering
    \includegraphics[height=1.5in]{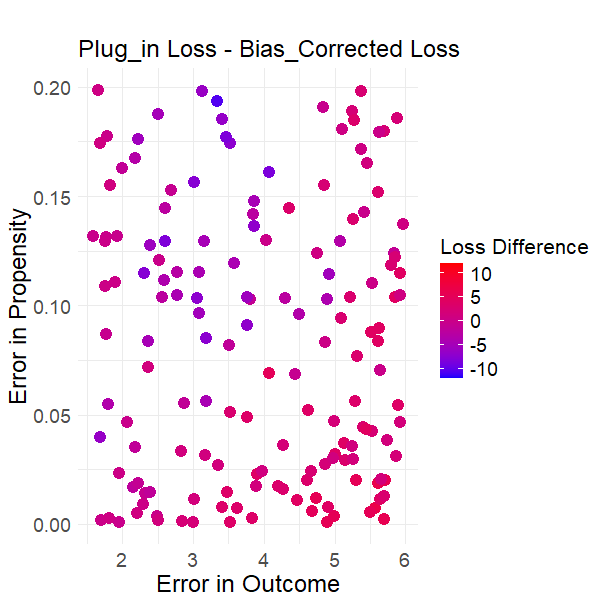}
    \includegraphics[height=1.4in]{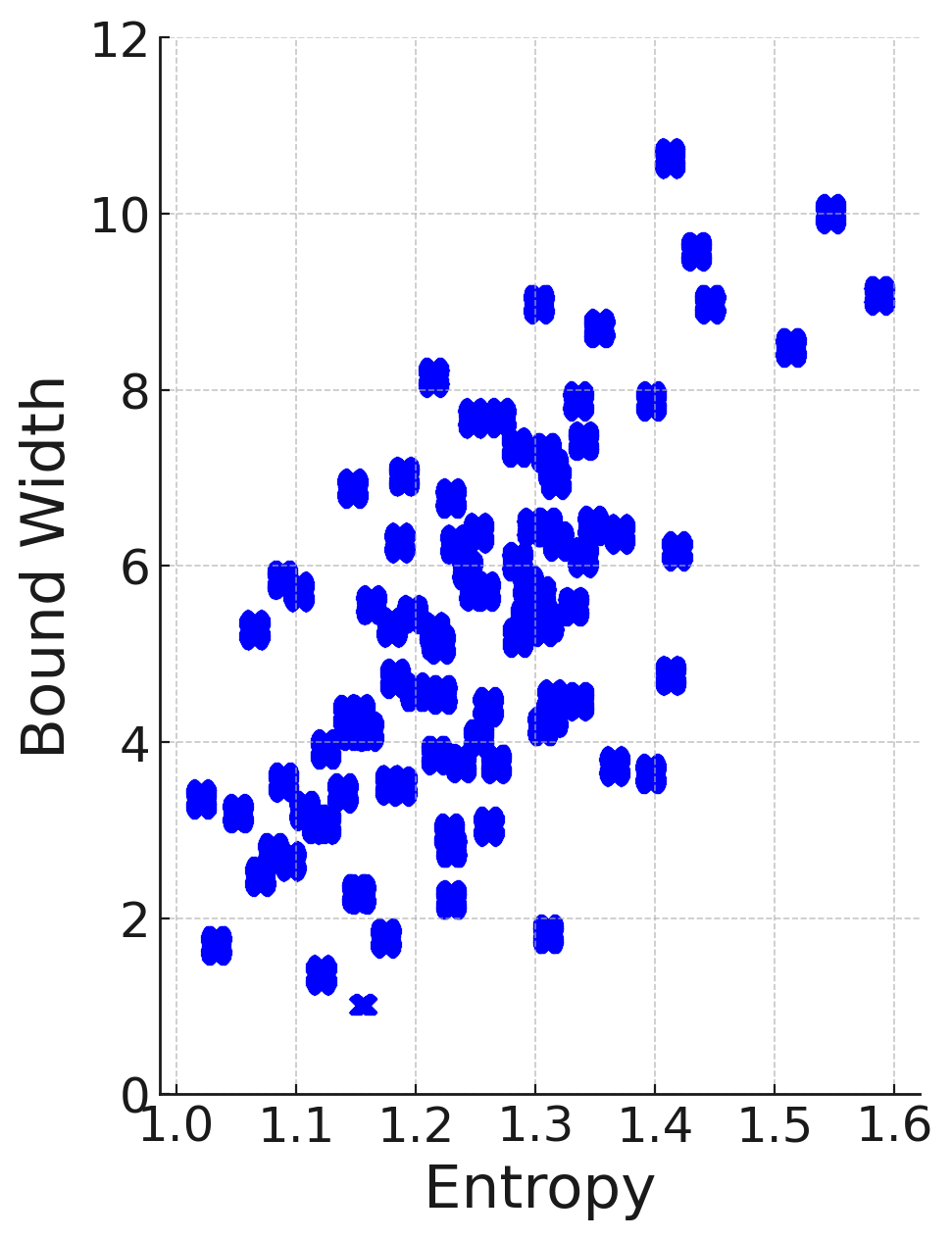}

    \caption{Top: difference in estimation error between the plug in and bias corrected estimators under varying errors in outcome and propensity modeling. Red is higher loss for the plug-in. All errors are measured in mean absolute deviation. Bottom: average worst-case bound width as a function of the entropy of $\nu$ where the range of outcomes is 40.}
    \label{fig:loss}
\end{figure}

\textbf{Results: }
 The left side of Figure \ref{fig:loss} demonstrates the difference in estimation error between the plug in and bias corrected estimators across 200 random seeds where we inject varying amounts of error into estimates of outcome and propensity models. We find that the bias-corrected model produces more accurate estimates of the bounds (known in simulation)  than the plug-in by large margins in the vast majority of situations, especially
when there is a high amount of error in the outcome regression. This is because the addition of the propensity-score modeling helps to correct for the bias in plug-in model. The plug-in model however does perform better when there is little error in outcome regression modeling and high error in propensity modeling. In this case, introducing the (poorly estimated) propensity score into the estimation adds more variance than it removes since the outcome models are already close to correct. However, we see strong performance from the bias-corrected estimator if either the outcome model has high error or the propensity model has low error, confirming that it provides strong performance over a wider range of the space.  

Intuitively, we might might expect $\nu$ ($\mathbb{P}(W|V=v,E=0)$), to be related to the width of our bounds based on the structure of $\gamma(v,w)$. To test this, we vary the distribution of $\nu$ and plot its entropy against the width of our bias-corrected bounds. We observe that as the entropy decreases, the average bound size also decreases (Figure \ref{fig:loss} right). This tells that as the known covariates in the study population become more predictive of the unaccounted for covariates in the target population, our bounds have better precision.

\begin{figure}[h!] 
    \centering
\includegraphics[height=1.5in]{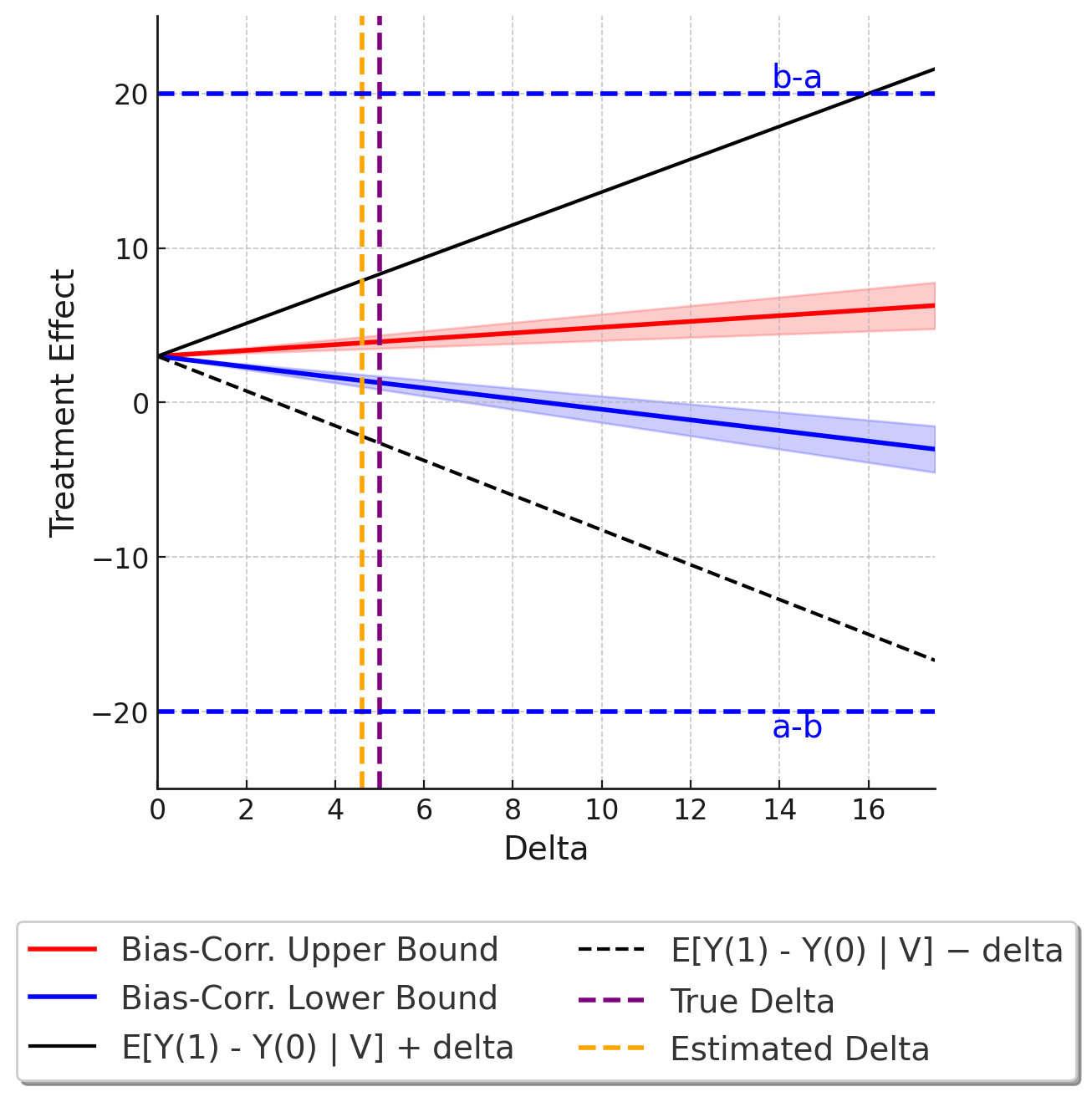}
\includegraphics[height=1.5in]{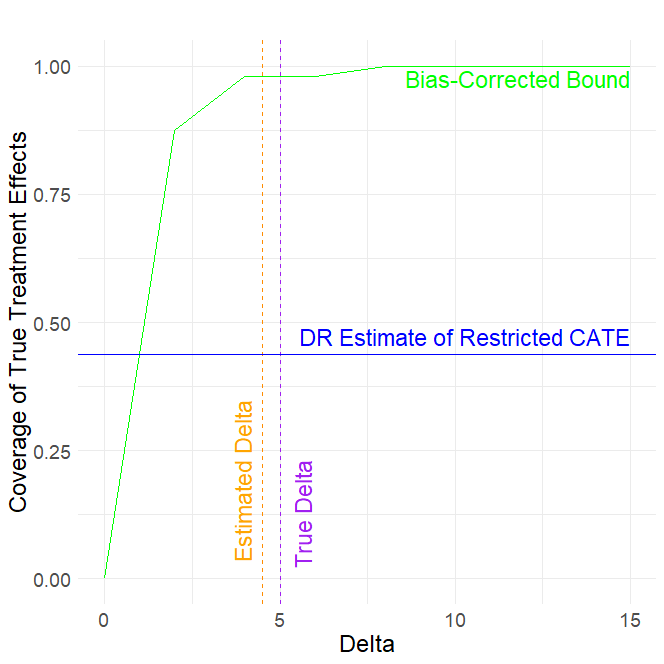}
    \caption{Simulation Data -- Top: bounds in the sensitivity model as a function of $\delta$, averaged over units. The red and blue lines are bounds output by our method (Bias Correction), with a 95$\%$ CI. The black lines are bounds that use only the sensitivity assumption in Equation \ref{eq:sensitivity} without our ecological inference framework. Bottom: Green line is the fraction of times our bounds cover the true CATE compared to blue line which is the coverage resulting 95$\%$ CI of the Restricted CATE   }
    \label{fig:sensitivity}
\end{figure}

\begin{figure}[h!] 
    \centering
\includegraphics[width=4.5cm,height=1.5in]{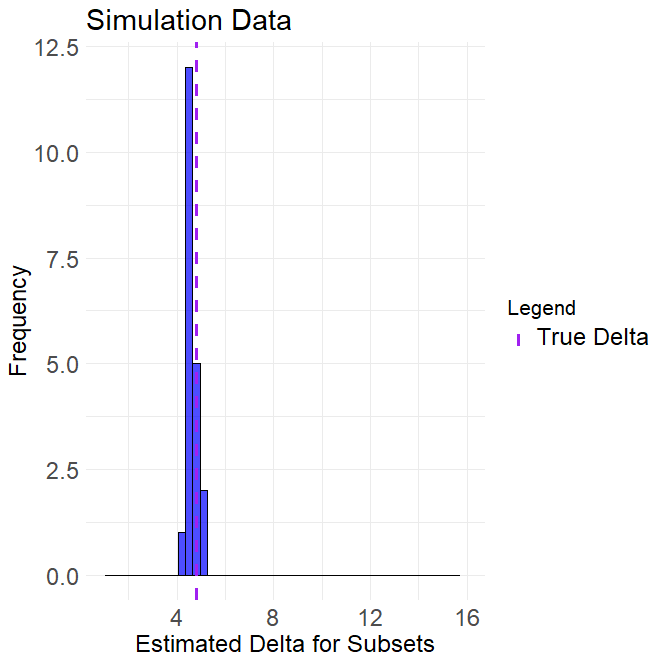}
\includegraphics[width=3.5cm, height=1.5in]{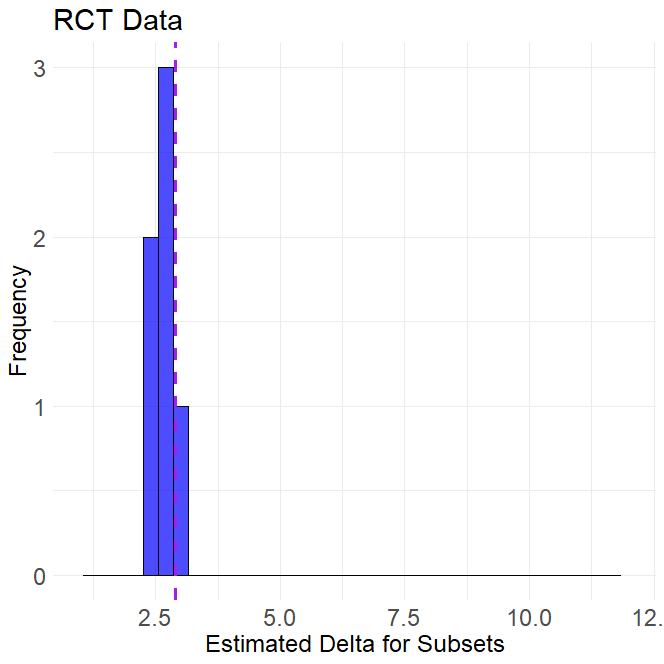}
    \caption{Benchmarking distribution for  Left: Simulation Data, Right: RCT Data. Final $\hat{\delta}$ represents the mean of each distribution and is within $\sim 10\%$ of both true $\delta$. We consider all potential subsets of V' and W' that can be made from V given the number of variables in W' (3 for Simulation and 1 for RCT)  }
    \label{fig:deltas}
    
\end{figure}

In Figure \ref{fig:sensitivity}, we apply the sensitivity model discussed in section \ref{section:sensitivity}. The left plot shows the average value of the bounds as a function of the sensitivity parameter $\delta$. As expected, the CATE is point-identified at $\delta = 0$, with progressively greater uncertainty as $\delta$ grows. For reference, we also plot bounds (in black) which use \textit{only} the sensitivity assumption in Equation \ref{eq:sensitivity}. By itself, this assumption implies bounds of width $2\delta$ for the CATE. The bounds output by our method are substantially stronger (less than half of the width), indicating that our ecological inference framework which uses the joint covariate distribution provides substantially more informative inferences than would otherwise be possible. The right figure shows the rate at which our bounds cover the true CATE as a function of $\delta$. As expected, if $\delta$ it close to its true value (known in the simulation), we observe close to nominal coverage levels. Additionally, Figure \ref{fig:sensitivity} shows that the 95$\%$ CI of the restricted CATE (measured on V and not W) baseline only correctly covers 43.7 percent of the true treatment effects, while at both the true and estimated values of our sensitivity parameter $\delta$ ($\hat{\delta}$ comes from Benchmarking distribution in Figure \ref{fig:deltas} (left), estimated and true $\delta$ are very close - within 10$\%$ of each other), roughly 98 percent of the true treatment effects are covered by our method. We can also observe that at both the true and estimated values of $\delta$, the bounds are very informative as the CI of the mean lower bound of the treatment effect ($\gamma_l(v,w)$) exceeds. More specifically, we have that $\gamma_l(v,w) \sim \in (1.1,1.6)$ at $\hat{\delta}$.


\begin{figure}
    \centering
    \includegraphics[height=1.5in]{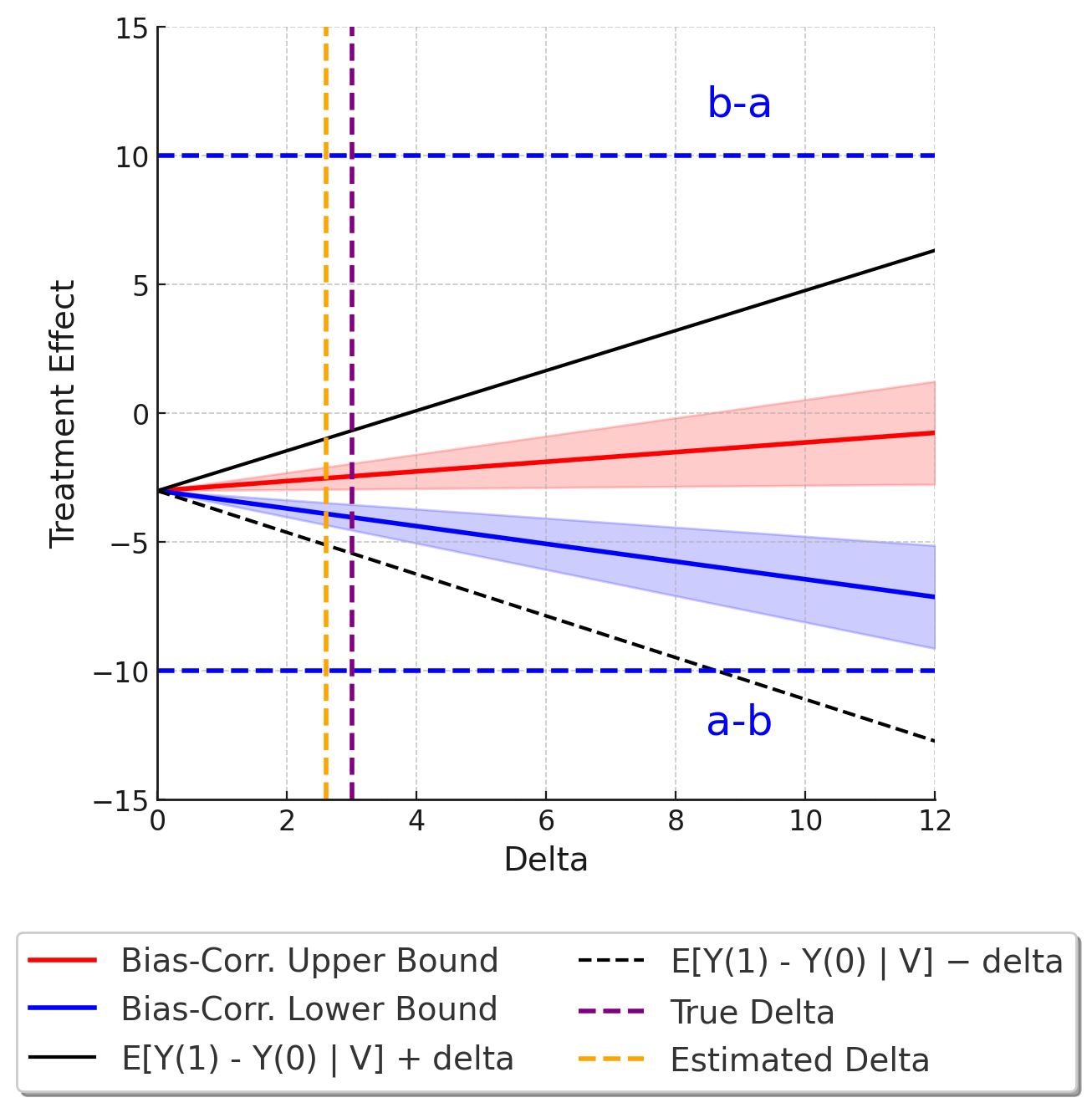}
\includegraphics[height=1.5in]{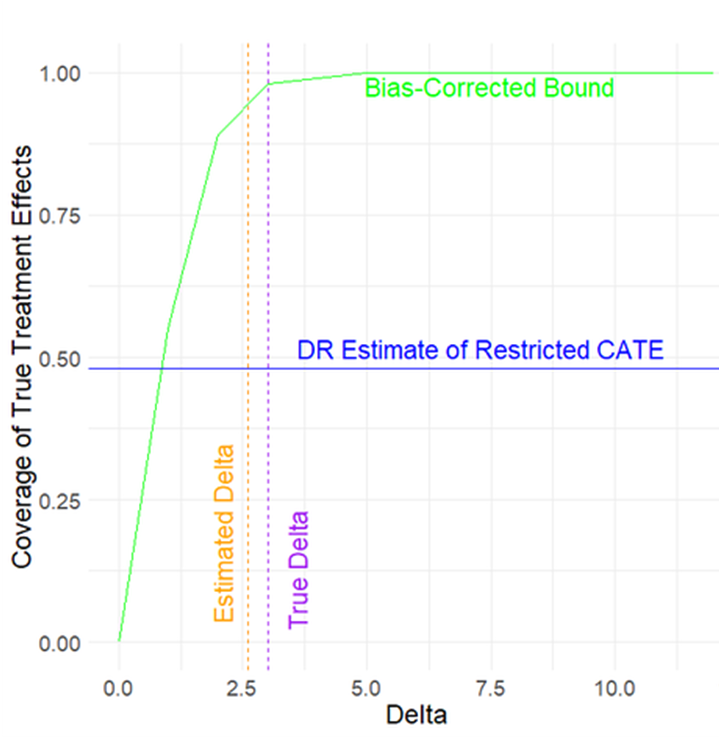}    \caption{RCT Data: labeling matches Figure \ref{fig:sensitivity}}
    \label{garging}
\end{figure}
Finally, we illustrate our method by estimating conditional treatment effects for a real-life RCT that measured the effect of a throat treatment (gargling with a licorice solution) prior to thoracic surgery on post-operative swallowing pain \citep{gargle}. We split the dataset into a ``study" and "target" population by holding out one of the covariates in the study population to form $W$. All of the variables in $V$ are standard measurements such as pain index, while the variable we separated for $W$ represents an indicator as to whether the patient experienced coughing after the ventilation tube was removed from their airway. Although this indicator is considered by many physicians to be a valuable metric (and would be visible to physicians working with the target population) it is frequently not recorded \citep{Duan_Zhang_Song_2021} and hence might not be available in a study dataset.  Further details about variables and the setting of the RCT can be found in the Appendix. Figure \ref{garging} shows the average width of bounds we obtain as a function of $\delta$, and we once again observe that the ecological framework provides substantially improved identifying power compared to naive bounds that only use the sensitivity assumption itself. We also provide the coverage levels of our estimator compared to the CI of the restricted CATE (although we can compare coverage only to an estimate of the fully conditional CATE, as the true function is not known for real data).

We again see very informative bounds in Figure \ref{garging}. In particular, the CI of the mean upper bound of the treatment effect lies entirely below 0 ($\gamma_u(v,w) \sim \in (-2.1,-2.8)$ at $\hat{\delta}$ where $\hat{\delta}$ comes from the Benchmarking process shown in Figure \ref{fig:deltas} right side, indicating a provable reduction in postoperative throat pain due to the treatment. At the estimated value of $\delta$, our bounds have almost exactly the desired 95\% coverage level. By contrast, the bounds output by a DR estimator for the restricted CATE have only 50\% coverage. This indicates that the held-out covariate has a significant impact on treatment effects, such that the expected effect for many patients moves outside the original CI after seeing the new covariate. Our bounds properly account for this uncertainty while remaining informative about effects. Similarly to the simulated data, even if we select a value of $\delta$ that matches the average width of CIs for the restricted CATE, our bounds have 25\% better coverage. This shows how leveraging the joint covariate distribution allows us to provide strictly more informative uncertainty quantification about treatment effects than otherwise would be possible.


\section{Discussion}
In this paper, we give the first formal presentation of an identification and estimation strategy for the CATE when generalizing to a target population that has covariates not observed in an earlier study. We develop a bias-corrected estimator that retains fast $O_{\mathbb{P}}(\frac{1}{\sqrt{n}})$ convergence rates even when nuisances are estimated nonparametrically, and is asymptotically normal under standard conditions. We also introduce a sensitivity model to bound impact of the new covariates on treatment effects. Empirically, we find that our method is often able to substantially reduce uncertainty about heterogeneous treatment effects, even in this challenging setting where no outcome data directly linked to the new covariates is observed. We caution that, like all causal inference methods, our framework requires domain expertise to assess the plausibility of assumptions, e.g. that $W$ and $V$ follow a consistent joint distribution between the populations, or the appropriate value of a sensitivity parameter $\delta$. However, when used appropriately, our framework gives users one way to assess the generalizability of effect estimates to newly identified subpopulations \textit{before} committing to a treatment assignment policy, helping to avert unintended negative consequences.  

\clearpage
\bibliography{biblio}
\newpage
\bibliographystyle{unsrtnat}
\newpage
\newpage
\onecolumn
\allowdisplaybreaks
\section{Appendix}
\subsection{Fully Conditional Bounds}
If we say that $Y \in [a,b]$ with probability one, then it would follow that
\begin{align*}
\E(Y \mid W=w, V=v) &= \frac{ \E(Y \mid V=v) -  \E(Y \mid W \neq w, V=v) \Pb(W \neq w \mid V=v)  }{ \Pb(W=w \mid V=v) } \\
&\in \bigg[ \max\left\{  \frac{ \E(Y \mid V=v) -  b \Pb(W \neq w \mid V=v)  }{ \Pb(W=w \mid V=v) } , a \right\} , \\
& \hspace{.5in}  \min\left\{  \frac{ \E(Y \mid V=v) -  a \Pb(W \neq w \mid V=v)  }{ \Pb(W=w \mid V=v) } , b \right\}  \bigg]
\end{align*}
Under the assumptions layed out in the Problem Setup: 
\begin{align*}
\E&(Y^1-Y^0 \mid V=v,W=w,E=1) \\
&=  \frac{ \E(Y^1-Y^0 \mid V=v, E=1) -  \E(Y^1-Y^0 \mid W \neq w, V=v,E=1) \Pb(W \neq w \mid V=v,E=0)  }{ \Pb(W=w \mid V=v, E=0) }   \\
&=  \frac{ \E(Y \mid V=v, A=1, E=1) - \E(Y \mid V=v, A=0, E=1) -  \alpha(v,w) \Pb(W \neq w \mid V=v,E=0)  }{ \Pb(W=w \mid V=v, E=0) }  
\end{align*}
where the first equality uses iterated expectation ,the second the fact that the treatment is randomized in the experiment, and defining $\alpha(w,v)=\E(Y^1-Y^0 \mid W \neq w, V=v,E=1)$. Now the only thing unknown is $\gamma(v,w)$. If all that is known is $Y \in [a,b]$ then $Y^1-Y^0 \in [a-b, b-a]$. We can then bound the conditional effects by simply utilizing the result from the bound on $\E(Y \mid W=w, V=v)$ and these worst-case values of $Y^1-Y^0 \in [a-b, b-a]$ for $\alpha(v,w)$. Doing so gives
$$ \gamma_\ell(v,w) \leq \E(Y^1-Y^0 \mid V=v,W=w,E=1) \leq \gamma_u(v,w) $$
where 
\begin{align*}
\gamma_\ell(v,w) &= \max\left\{ \frac{ \mu_1(v) - \mu_0(v)  - (b-a) \Pb(W \neq w \mid V=v,E=0)  }{ \Pb(W=w \mid V=v, E=0) }  , (a-b) \right\} \\
\gamma_u(v,w) &= \min\left\{ \frac{ \mu_1(v) - \mu_0(v)  - (a-b) \Pb(W \neq w \mid V=v,E=0)  }{ \Pb(W=w \mid V=v, E=0) }  , (b-a) \right\}
\end{align*}
and where we have defined
$$ \mu_a(v) = \E(Y \mid V=v, A=a, E=1) . $$
\subsection{Moment Condition}
These steps show how to go from (6) to (7) in section 4.2 \\
Let $\gamma_\ell(v,w) = \max\{\tau_\ell(x),a-b\}$. Then 
\begin{align*}
    M(\beta)  &= E [g(x)(\gamma(v,w)-m(v, w; \beta))] \\
    &= E [g(x)(\max\{\tau_\ell(x),a-b\}-m(v, w; \beta))] \\
    &= E[g(x) \{\tau_\ell(x)* 1(\tau_\ell(x) \geq (a-b)) + (a-b)* 1(\tau_\ell(x) \leq (a-b)) -m(v, w; \beta)  \}] \\
    &= E[g(x)\{(\tau_\ell(x)+b-a)*1(\tau_\ell(x)+b-a \geq 0)+a-b -m(v, w; \beta) \}  ]
\end{align*}
\clearpage
Let $\gamma_u(v,w) = \min\{\tau_u(x),b-a\}$. Then 
\begin{align*}
    M(\beta)  &= E [g(x)(\gamma(v,w)-m(v, w; \beta))] \\
    &= E [g(x)(\min\{\tau_u(x),b-a\}-m(v, w; \beta))] \\
    &= E[g(x)\{(\tau_u(x)+a-b)*1(\tau_u(x)+a-b \leq 0)+b-a -m(v, w; \beta) \}  ]
\end{align*}
\subsection{Use of Margin Condition}
\begin{align*}
    E[\tau \mathbb{1}(\hat{\tau}+b-a) -\tau \mathbb{1}(\tau+b-a) ] 
    &= E[\tau \mathbb{1}(\hat{\tau}) -\tau \mathbb{1}(\tau) ] \\
    & \leq E[|\tau| \cdot | \mathbb{1}(\hat{\tau}) -\mathbb{1}(\tau)|  ] \\
    & \leq \int |\hat{\tau}-\tau| \mathbb{1}(|\tau|\leq |\hat{\tau}-\tau|) dP(x) \\
    & \leq \int |\hat{\tau}-\tau| \mathbb{1}(|\tau|\leq |\hat{\tau}-\tau|) dP(x) \\
    & \leq \sup_x |\hat{\tau}-\tau| \cdot P(|\tau|\leq ||\hat{\tau}-\tau||_{\infty})  \\
    & \leq C ||\hat{\tau}-\tau||_{\infty}^{1+\alpha} \\ 
\end{align*}
The third line follows from the fact that if $\tau$ and $\hat{\tau}$ are of opposite signs,then $\tau|\leq |\hat{\tau}-\tau|$.
\subsection{Influence Function Derivation}

Facts: These are all known statistical facts and influence functions. We use the strategy of using known influence functions to construct construct more complicated influence functions (IF Calculus) detailed in section 3.4.3 of \cite{kennedy2023semiparametric} (This section and paper in general is a great resource for understanding influence functions). It is important to note that we use the IF Calculus as a heuristic to derive the Influence Function. In Theorem 3, we show that the bias is second order. In doing this, we have implicitly shown that our estimator satisfies a Von-Mises expansion. \citep{kennedy2023semiparametric} This shows we have a valid Influence Function, and the fact that we are using a non-parametric model means that we have correctly identified the only Influence Function for the bounds in Theorem 1.

\begin{align*}
    IF(E(Y|X=x)) &= \frac{1(X=x)}{P(X=x)}(Y-E[Y|X=x]) \\
    P(W=w|X=x) &= E[1(W=w)|X=x] \\
    \frac{p(x)}{p(V=v,A=1,E=1)} &= \frac{p(W=w|V=v)}{p(A=1,E=1|V=v)} \\
    IF (p(x)) &= \{1(X=x)-p(x)\}
\end{align*}
Derivation:
Here we say $\tau_n$ is the numerator of $\tau$ or $$\mu_1(v) - \mu_0(v)  - (b-a) \Pb(W \neq w \mid V=v,E=0)$$ and $\tau_d$ is the denominator $\tau$ or $\Pb(W=w \mid V=v, E=0)$. As mentioned in the paper $f(x) = 1(\tau_\ell(x)+b-a \geq 0)$. 
\begin{align*}
    IF(\tau(x)) &= IF(\frac{\tau_n(x)}{\tau_d(x)}) \\
    &= \frac{IF(\tau_n(x))}{\tau_d(x)} - \frac{\tau_n(x)}{\tau_d(x)^2} IF(\tau_d(x)) \\
    IF(\tau_n(x)) &= \frac{EA *1(V=v)}{p(V=v,A=1,E=1)}(Y-\mu_1(v)) - \frac{E(1-A)*1(V=v)}{p(V=v,A=0,E=1)}(Y-\mu_0(v))\\
    & + \frac{(b-a) 1(V=v,E=0)}{p(V=v,E=0)} (1 (W = w) - p(W = w | V=v, E=0) \\
    IF(\tau_d(x)) &= \frac{1(V=v, E=0 )}{P(V=v,E=0)} (1(W=w) - P(W=w | V=v,E=0))
\end{align*}

 We call the known portion q(x) such that $q(x) = -((b-a)f(x) +a-b -m(x; \beta))$ . We then have
\begin{align*}
    & IF \big [\sum_x g(x)\{\tau(x)f(x)-q(x) \} p(x)\big]  \\
    &=  \big [\sum_x ( g(x)\{IF(\tau(x))f(x) \} p(x) +g(x)\{\tau(x)f(x)-q(x) \} IF(p(x)))\big]  \\
    &=  \big [\sum_x (g(x) \{ \frac{IF(\tau_n(x))}{\tau_d(x)} - \frac{\tau(x)IF(\tau_d(x))}{\tau_d(x)} \}
    f(x)p(x))
    +g(X)\{\tau(X)f(X)-q(X)\} \big]  \\
\end{align*}
Let 
\begin{align*}
    \varphi_1(X;\beta) &= \sum_x g(x) \{ \frac{IF(\tau_n(x))}{\tau_d(x)} \}
    f(x)p(x) \\
    \varphi_2(X;\beta) &= -\sum_x g(x) \{  \frac{\tau(x)IF(\tau_d(x))}{\tau_d(x)} \}
    f(x)p(x) \\
    \varphi_3(X;\beta) &= g(x)\{\tau(x)f(x)-q(x)\} \\
\end{align*}

Lets further examine the first two pieces.
\begin{align*}
    \varphi_1(X;\beta) &=  \frac{EA}{p(A=1,E=1|V)}(Y-\hat{\mu}_1(V))
    *\sum_w g(V,w) f(V,w) p(w|V)*\frac{1}{\hat{\tau}_d(V,w)}
    \\
    & - \frac{E(1-A)}{p(A=0,E=1|V)}(Y-\hat{\mu}_0(V))*\sum_w g(V,w) f(V,w) p(w|V)*\frac{1}{\hat{\tau}_d(V,w)} \\
&+ \frac{(b-a)(1-E)}{p(E=0|V)} \{ 
g(V,W) f(V,W) *\frac{p(W|V)}{\hat{\tau}_d(V,W)} 
\\& - \sum_w g(V,w) f(V,w) p(w|V)*\frac{p(W=w | V, E=0)}{\hat{\tau}_d(V,w)} 
\} \\
&= \frac{EA}{p(A=1,E=1|V)}(Y-\hat{\mu}_1(V))
    *\sum_w g(V,w) f(V,w) 
    \\
    & - \frac{E(1-A)}{p(A=0,E=1|V)}(Y-\hat{\mu}_0(V))*\sum_w g(V,w) f(V,w) \\
&+ \frac{(b-a)(1-E)}{p(E=0|V)} \{ 
g(V,W) f(V,W)  - \sum_w g(V,w) f(V,w) p(w|V)
\}
\end{align*}
To go from the first equality to the second - we use the fact that $\hat{\tau}_d(V,W) = \hat{p}(W|V)$ (it does not matter if E=0 or E=1 for $\hat{p}(W|V)$ under the assumptions laid out in the Problem Setup)
\begin{align*}
    \varphi_2(X;\beta) &=  - \frac{1-E}{\hat{p}(E=0|V)} \{   
g(V,W) f(V,W) \hat{p}(W|V)*\frac{\hat{\tau(X)}}{\hat{\tau}_d(X)}  \\ &- \sum_w g(V,w) f(V,w) \hat{p}(w|V)*\frac{\hat{\tau}(V,w)\hat{p}(W=w | V, E=0)}{\hat{\tau}_d(V,w)}
\} \\
&= - \frac{1-E}{\hat{p}(E=0|V)} \{   
g(V,W) f(V,W) \hat{\tau}(X) \\
&- \sum_w g(V,w) f(V,w) \hat{\tau}(V,w)\hat{p}(W=w | V, E=0) \}
\end{align*}
Ultimately, we have
$$\varphi(X;\beta)  = \varphi_1 (X; \beta) +  \varphi_2 (X; \beta) + \varphi_3 (X; \beta)$$
and the result given in the paper follows as $g(x) = \begin{bmatrix}
v \\
w 
\end{bmatrix} = x $

\begin{align*}
    \varphi(X;\beta)  &= \frac{EA}{p(A=1,E=1|V)}(Y-\hat{\mu}_1(V))
    *\sum_w g(V,w) f(V,w) 
    \\
    & - \frac{E(1-A)}{p(A=0,E=1|V)}(Y-\hat{\mu}_0(V))*\sum_w g(V,w) f(V,w) \\
&+ \frac{(b-a)(1-E)}{p(E=0|V)} \{ 
g(V,W) f(V,W)  - \sum_w g(V,w) f(V,w) p(w|V) \\
&- \frac{1-E}{\hat{p}(E=0|V)} \{   
g(V,W) f(V,W) \hat{\tau}(X) \\
&- \sum_w g(V,w) f(V,w) \hat{\tau}(V,w)\hat{p}(W=w | V, E=0) \}\\
&+ g(X) \{\tau_\ell(X)f(X)+(b-a)f (X)+a-b - m(X,\beta) \} 
\end{align*}
The influence function for $\gamma_u(x)$ would be 
\begin{align*}
    \varphi(X;\beta)  &= \frac{EA}{p(A=1,E=1|V)}(Y-\hat{\mu}_1(V))
    *\sum_w g(V,w) f(V,w) 
    \\
    & - \frac{E(1-A)}{p(A=0,E=1|V)}(Y-\hat{\mu}_0(V))*\sum_w g(V,w) f(V,w) \\
&+ \frac{(b-a)(1-E)}{p(E=0|V)} \{ 
g(V,W) f(V,W)  - \sum_w g(V,w) f(V,w) p(w|V) \\
&- \frac{1-E}{\hat{p}(E=0|V)} \{   
g(V,W) f(V,W) \hat{\tau}(X) \\
&- \sum_w g(V,w) f(V,w) \hat{\tau}(V,w)\hat{p}(W=w | V, E=0) \}\\
&+ g(X) \{\tau_\ell(X)f(X)+(a-b)f (X)+a-b - m(X,\beta) \} 
\end{align*}
For the upper bound for the influence function and going forward the only difference from the lower bound is that the indicator becomes $f(X) = 1(\tau_u(x)+a-b \leq 0)$ and that a-b and b-a are flipped wherever they occur
\subsubsection{Bias Derivation}
In the steps below, we primarily use iterated expectation and algebraic manipulation - for notation purposes we say $P(\cancel{W}|V) = 1- P(W|V)$. We will end up converting some of the nuisances to the notation we use in the paper (eg $\nu(V,w) =p(W=w| V, E=0) $,$\rho_0(V)= p(E=0|V)$) but for now we leave them in their original form as it makes them easier to visualize and manipulate. For brevity though we do use $\pi_x  = p(A=x,E=1|V)$ in these proofs. It is also important to be clear again that 
$\hat{\tau}_d(V,W) = \hat{p}(W|V)$ so we can interchange these as we like. 
Iterated expectation is the most common technique used . Our strategy will be to create second order terms wherever we can and then collect the remainder. We focus first on turning each first order term into a first order term and a second order term. The remaining first order terms will be dealt with after. The bias is given by:
\begin{align*}
    R_n = \mathbb{P}\{\varphi(X; \beta, \hat{\eta}) - \varphi(X; \beta, \eta_0)\} = \mathbb{P}\{\varphi(X; \beta, \hat{\eta})\}
\end{align*}
We then have
\begin{align*}
    & \mathbb{E}\{\varphi_1(X; \beta, \hat{\eta})\} = 
    \mathbb{E}\big [ \frac{EA}{\hat{p}(A=1,E=1|V)}(Y-\hat{\mu}_1(v))
    *\sum_x g(V,w) f(V,w) 
    \\
    & - \frac{E(1-A)}{\hat{p}(A=0,E=1|V)}(Y-\hat{\mu}_0(v))*\sum_x g(V,w) f(V,w) \ \\
&+ \frac{(b-a)(1-E)}{\hat{p}(E=0|V)} \{ 
g(X) f(X)  - \sum_w g(V,w) f(V,w) \hat{p}(w|V) 
\} \big] \\
&= 
    \mathbb{E}\big [ \frac{\pi_1}{\hat{\pi}_1}(\mu_1(V)-\hat{\mu}_1(V))
    *\sum_x g(V,w) f(V,w) 
    \\
    & -  \frac{\pi_0}{\hat{\pi}_0}(\mu_0(V)-\hat{\mu}_0(V))*\sum_x g(V,w) f(V,w)  \\
&+ \frac{(b-a)p(E=0|X)}{\hat{p}(E=0|V)} 
g(X) f(X)  \\
&- \frac{(b-a)p(E=0|V)}{\hat{p}(E=0|V)}\sum_w g(V,w) f(V,w) \hat{p}(w|V) 
 \big] \\
&= 
\mathbb{E}\big [ \frac{\pi_1-\hat{\pi}_1}{\hat{\pi}_1}(\mu_1(V)-\hat{\mu}_1(V))
    *\sum_x g(V,w) f(V,w) 
    \\
    & -  \frac{\pi_0-\hat{\pi}_0}{\hat{\pi}_0}(\mu_0(V)-\hat{\mu}_0(V))*\sum_x g(V,w) f(V,w) \hat{p}(w|V) \\
&+ (\mu_1(V)-\hat{\mu}_1(V))
    *\sum_x g(V,w) f(V,w) \ \\
&-(\mu_0(V)-\hat{\mu}_0(V))*\sum_x g(V,w) f(V,w)  \\
&+ 
\frac{(b-a)p(E=0|X)}{\hat{p}(E=0|V)} 
g(X) f(X)  
- \frac{(b-a)p(E=0|V)}{\hat{p}(E=0|V)}\sum_w g(V,w) f(V,w) \hat{p}(w|V) 
\big]
\end{align*}
The last line is slightly more tricky to get a second order term out of - let's focus on that now:
\begin{align*}
     &\mathbb{E} \big[ (b-a)\sum_x \frac{p(E=0|x)}{\hat{p}(E=0|v)} 
g(x) f(x)p(w|v)p(v) \big]
\\&- (b-a)\sum_x  \frac{p(E=0|v)}{\hat{p}(E=0|v)} g(x) f(x) \hat{p}(w|v)p(V) \\
&=\mathbb{E} \big[ (b-a)\sum_x \frac{p(E=0|x)}{\hat{p}(E=0|v)} 
g(x) f(x)p(w|v)p(v)
\\& - (b-a)\sum_x \frac{p(E=0|v)}{\hat{p}(E=0|v)} g(x) f(x) \hat{p}(w|v)p(v)\big] \\
&=\mathbb{E} \big[(b-a)\sum_x g(x) f(x)p(v) \{ \frac{p(w|v)p(E=0|x)}{\hat{p}(E=0|v)} - \frac{\hat{p}(W=w | V, E=0)p(E=0|v)}{\hat{p}(E=0|v)}   \} \big]\\
&=\mathbb{E} \big[ (b-a)\sum_x g(x) f(x)p(v)
\{ \frac{p(E=0|v)}{\hat{p}(E=0|v)}-1   \}*
\\&
\{  p(W=w | V=v, E=0)- \hat{p}(W=w | V=v, E=0)  \}
\\& +(b-a)\sum_x g(x) f(x)
\{  p(W=w | V=v, E=0)- \hat{p}(W=w | V=v, E=0)  \}\big] \\
&=\mathbb{E} \big[ (b-a)\sum_w g(V,w) f(V,w)
\{ \frac{p(E=0|V)}{\hat{p}(E=0|V)}-1   \}
\{  p(W=w | V, E=0)- \hat{p}(W=w | V, E=0)  \}\big] 
\\&a+\mathbb{E} \big[ (b-a)\sum_w g(V,w) f(V,w)
\{  p(W=w | V, E=0)- \hat{p}(W=w | V, E=0)  \} \big]\\
\end{align*}
Now let's handle the remaining first order terms in $\varphi_1$ and the rest of $\varphi_2$ and $\varphi_3$. We can decompose the components of $\varphi$ into (remember that $\varphi_3 = g(x)\{\tau(x)f(x)-q(x)\}$):
\begin{align*}
    \varphi &= \varphi_1(X;\beta) + \varphi_2(X;\beta)   + g(X)( \frac{\hat{\tau}_n(X)}{\hat{\tau}_d(X)}f -q(X)) - g(X)( \frac{\tau_n}{\tau_d(X)}f -q(X)) \\
    &= \varphi_2(X;\beta)    + g(X)( \frac{\hat{\tau}_n(X)}{\hat{\tau}_d}f -q(X))- g(X)( \frac{\hat{\tau}_n(X)}{\tau_d}f -q(X)) \\
    &+ \varphi_1(X;\beta)  +g(X)( \frac{\hat{\tau}_n(X)}{\tau_d(X)}f -q(X)) - g(X)( \frac{\tau_n(X)}{\tau_d(X)}f -q(X))
\end{align*}
Lets examine the first line of the decomposition
\begin{align*}
     &\varphi_2   + g( \frac{\hat{\tau}_n}{\hat{\tau}_d}f(X) -q(X))- g( \frac{\hat{\tau}_n}{\tau_d}f(X) -q(X)) \\
     &= \mathbb{E} \big[  -\frac{1-E}{p(E=0|V)} \{   
g(X) f(X)(X) p(w|v)*\frac{\hat{\tau}(X)}{\tau_d(X)} + \\& \sum_w g(V,w) f(X)(V,w) p(w|V)*\frac{\hat{\tau}(V,w)p(W=w | V, E=0)}{\tau_d(V,w)}
\}  
\\ & + g( \frac{\hat{\tau}_n}{\hat{\tau}_d}f(X) -q(X))- g( \frac{\hat{\tau}_n}{\tau_d}f(X) -q(X)) \big] 
\\ &= \mathbb{E} \big[ -\frac{p(E=0|X)}{\hat{p}(E=0|V)}  
g(X) f(X)(X) \frac{\hat{\tau}_n(X)}{\hat{\tau}_d(X)}  \\&+  \frac{p(E=0|V)}{\hat{p}(E=0|V)}\sum_w \big ( g(V,w) f(V,w) \hat{p}(w|V)*\frac{\hat{\tau}_n(V,w)}{\hat{\tau}_d(V,w)} \big )
+ gf( \frac{\hat{\tau}_n}{\hat{\tau}_d}- \frac{\hat{\tau}_n}{\tau_d} ) \big]  \\
&= \mathbb{E} \big[
 \sum_w \frac{\hat{\tau}_n g(V,w) f(V,w)}{\hat{\tau}_d(V,w)} 
\{ \frac{p(E=0|V)}{\hat{p}(E=0|V)}-1   \}
\{   \hat{p}(W=w | V, E=0)-p(W=w | V, E=0)  \}+ 
\\& \sum_w \big( \frac{\hat{\tau}_n g(V,w) f(V,w)}{\hat{\tau}_d(V,w))} 
\{  \hat{p}(W=w | V, E=0) - p(W=w | V, E=0)  \} \big )
+ gf( \frac{\hat{\tau}_n}{\hat{\tau}_d}- \frac{\hat{\tau}_n}{\tau_d} ) \big]
\end{align*}
Lets examine the second line of the last expectation as the first line is already second order
\begin{align*}
    & \mathbb{E} \big[  \sum_w \big( \frac{\hat{\tau}_n g(V,w) f(V,w)}{\hat{\tau}_d(V,w)} 
\{  \hat{p}(W=w | V, E=0)-p(W=w | V, E=0)  \} \big ) 
+ gf( \frac{\hat{\tau}_n}{\hat{\tau}_d}- \frac{\hat{\tau}_n}{\tau_d} ) \big] \\
&=  \mathbb{E} \big[  \sum_w \big (\frac{\hat{\tau}_n g(V,w) f(V,w)}{\hat{\tau}_d(V,w)} 
\{    \hat{\tau}_d (V,w)-\tau_d (V,w)   \} \big )
+ gf\hat{\tau}_n( \frac{1}{\hat{\tau}_d}- \frac{1}{\tau_d} ) \big] \\
&=  \mathbb{E} \big[  \sum_w \frac{\hat{\tau}_n g(V,w) f(V,w)}{\hat{\tau}_d(V,w)} 
\{  \hat{\tau}_d - \tau_d   \} 
+ \hat{\tau}_n g(V,w) f(V,w)\hat{p}(w|V)
( \frac{1}{\hat{\tau}_d}- \frac{1}{\tau_d} ) \big] \\
&= \mathbb{E} \big[  \sum_w \hat{\tau}_n g(V,w) f(V,w) \big( \frac{1}{\hat{\tau}_d(V,w)} 
\{    \hat{\tau}_d(V,w)-\tau_d(V,w)  \} 
+ 
( \frac{1}{\hat{\tau}_d(V,w)}- \frac{1}{\tau_d(V,w)} ) \big) \big] \\
&= \mathbb{E} \big[  \sum_w \hat{\tau}_n g(V,w) f(V,w)
\big(
  -\frac{\tau_d(V,w)}{\hat{\tau}_d(V,w)}  + \frac{1}{\hat{\tau}_d(V,w)}
+ 
\frac{1}{\hat{\tau}_d(V,w)}- \frac{1}{\tau_d(V,w)}  \big) \big] \\
&= \mathbb{E} \big[  \sum_w \hat{\tau}_n g(V,w) f(V,w)* \frac{1}{\tau_d(V,w) \hat{\tau}_d(V,w) }
\big( \hat{\tau}_d(V,w)-\tau_d(V,w)
    \big)^2 \big] \\
\end{align*}
Now let's focus on: $$\varphi_1(X;\beta)  +g(X)( \frac{\hat{\tau}_n(X)}{\tau_d(X)}f -q(X)) - g(X)( \frac{\tau_n(X)}{\tau_d(X)}f -q(X))$$ 
\begin{align*}
    & \varphi_1 +g( \frac{\hat{\tau}_n}{\tau_d}f -q(X)) - g( \frac{\tau_n}{\tau_d}f -q(X)) \\
    &= \mathbb{E}\big [ \frac{\pi_1-\hat{\pi}_1}{\hat{\pi}_1}(\mu_1(V)-\hat{\mu}_1(V))
    *\sum_x g(V,w) f(V,w) 
    \\
    & -  \frac{\pi_0-\hat{\pi}_0}{\hat{\pi}_0}(\mu_0(V)-\hat{\mu}_0(V))*\sum_x g(V,w) f(V,w)  \\
&+ (b-a)\sum_w g(V,w) f(V,w)
\{ \frac{p(E=0|V)}{\hat{p}(E=0|V)}-1   \}
\{  p(W=w | V, E=0)- \hat{p}(W=w | V, E=0)  \} \\
&+ (\mu_1(V)-\hat{\mu}_1(V))
    *\sum_x g(V,w) f(V,w)  \\
&-(\mu_0(V)-\hat{\mu}_0(V))*\sum_x g(V,w) f(V,w)  \\
&+ (b-a)\sum_w g(V,w) f(V,w)
\{  p(W=w | V, E=0)- \hat{p}(W=w | V, E=0)  \}
\\
& +g(X)( \frac{\hat{\tau}_n(X)}{\tau_d(X)}f -q(X)) - g( \frac{\tau_n(X)}{\tau_d(X)}f -q(X))
\big]
\end{align*}
The first 3 lines of the expectation are already second order so let's examine the rest - we have:
\begin{align*}
 \mathbb{E}\big [
&(\mu_1(V)-\hat{\mu}_1(V))
    *\sum_w g(V,w) f(V,w)  \\
&-(\mu_0(V)-\hat{\mu}_0(V))*\sum_w g(V,w) f(V,w)  \\
&+ (b-a)\sum_w g(V,w) f(V,w)
\{  p(W=w | V, E=0)- \hat{p}(W=w | V, E=0)  \}
\\
& +g(X)f(X) (\frac{\hat{\mu}_1(V) - \hat{\mu}_0(V)  - (b-a) \hat{\Pb}(\cancel{W}  \mid V,E=0)}{\tau_d(X)} - \frac{\mu_1(V) - \mu_0(V)  - (b-a) \Pb(\cancel{W} \mid V,E=0)}{\tau_d(X)})
\big] \\
&=  \mathbb{E}\big [
\frac{\mu_1(V)-\hat{\mu}_1(V)}{\hat{\tau}_d(X)}
    g(X) f(X)  \\
&-\frac{\mu_0(V)-\hat{\mu}_0(V)}{\hat{\tau}_d(X)}
    g(X) f(X)  \\
&+ (b-a)\frac{ g(X) f(X)
\{  p(W=w | V, E=0)- \hat{p}(W=w | V, E=0)  \}}{\hat{\tau}_d(X)}
\\
& +g(X)f(X) (\frac{\hat{\mu}_1(V) - \hat{\mu}_0(V)  - (b-a) \hat{\Pb}(\cancel{W}  \mid V,E=0)}{\tau_d(X)} - \frac{\mu_1(V) - \mu_0(V)  - (b-a) \Pb(\cancel{W} \mid V,E=0)}{\tau_d(X)})
\big] \\
&=  \mathbb{E}\big [g(V,w)f(X)(\frac{\mu_1(V)}{\hat{\tau}_d(X)}-\frac{\hat{\mu}_1(V)}{\hat{\tau}_d(X)}+\frac{\hat{\mu}_1(v)}{\tau_d(X)} - \frac{\mu_1(v)}{\tau_d(X)})
    \\
&-g(V,w)f(X)(\frac{\mu_0(V)}{\hat{\tau}_d(X}-\frac{\hat{\mu}_0(V)}{\hat{\tau}_d(X)}+\frac{\hat{\mu}_0(v)}{\tau_d(X)} - \frac{\mu_0(v)}{\tau_d(X)})
    \\
&+ (b-a)\sum_w g(V,w) f(V,w)
\{ (1- p(W \neq w | V, E=0))- (1-\hat{p}(W \neq w | V, E=0))  \}
\\
& +g(X)f(X) (\frac{  - (b-a) \hat{\Pb}( \cancel{W} \mid V,E=0)}{\tau_d(X)} - \frac{ - (b-a) \Pb( \cancel{W} \mid V,E=0)}{\tau_d(X)})
\big] \\
&=  \mathbb{E}\big [g(X)f(X) \{ (\mu_1(V)-\hat{\mu}_1(V))(\frac{1}{\hat{\tau}_d(X)}- \frac{1}{\tau_d(X)}) \}
    \\
&-g(X)f(X) \{ (\mu_0(V)-\hat{\mu}_0(V))(\frac{1}{\hat{\tau}_d(X)}- \frac{1}{\tau_d(X)}) \}
    \\
&- (b-a) g(X) f(X)
 (\frac{p(\cancel{W} \mid V, E=0)}{\hat{\tau}_d(X)} -\frac{\hat{p}(\cancel{W} \mid V, E=0)}{\hat{\tau}_d(X)} 
 \\ & +\frac{\hat{p}(\cancel{W} \mid V, E=0)}{\tau_d(X)} - \frac{p(\cancel{W} \mid V, E=0)}{\tau_d(X)}  )
\big] \\
&=  \mathbb{E}\big [g(X)f(X) \{ (\mu_1(V)-\hat{\mu}_1(V))(\frac{1}{\hat{\tau}_d(X)}- \frac{1}{\tau_d(X)}) \}
    \\
&-g(X)f(X) \{ (\mu_0(V)-\hat{\mu}_0(V))(\frac{1}{\hat{\tau}_d(X)}- \frac{1}{\tau_d(X)}) \}
    \\
&- (b-a) g(X) f(X)
 (p(\cancel{W} \mid V, E=0) - \hat{p}(\cancel{W} \mid V, E=0))({\hat{\tau}_d(X)}- \frac{1}{\tau_d(X)}) 
\big] \\
&=  \mathbb{E}\big [g(X)f(X) \{ (\mu_1(V)-\hat{\mu}_1(V))(\frac{1}{\hat{\tau}_d(X)}- \frac{1}{\tau_d(X)}) \}
    \\
&-g(X)f(X) \{ (\mu_0(V)-\hat{\mu}_0(V))(\frac{1}{\hat{\tau}_d(X)}- \frac{1}{\tau_d(X)}) \}
    \\
&- (b-a) g(X) f(X)
 (\hat{p}(W \mid V, E=0) -p(W \mid V, E=0))(\frac{1}{\hat{\tau}_d(X)}- \frac{1}{\tau_d(X)}) 
\big] \\
\end{align*}

Final Bias (from collecting all the pieces above)
\begin{align*}
    & \mathbb{E} \big[
 \sum_w \frac{\hat{\tau}_n g(V,w) f(V,w)}{\hat{\tau}_d(V,w)}
\{ \frac{p(E=0|V)}{\hat{p}(E=0|V)}-1   \}
\{   \hat{p}(W=w | V, E=0)-p(W=w | V, E=0)  \}+ 
\\& \sum_w \hat{\tau}_n g(V,w) f(V,w)* \frac{1}{ \hat{\tau}_d(V,w)^2 }
\big( \hat{\tau}_d(V,w)-\tau_d(V,w)
    \big)^2
    +\\&
    \frac{\pi_1-\hat{\pi}_1}{\hat{\pi}_1}(\mu_1(V)-\hat{\mu}_1(V))
    *\sum_x g(V,w) f(V,w) 
    \\
    & -  \frac{\pi_0-\hat{\pi}_0}{\hat{\pi}_0}(\mu_0(V)-\hat{\mu}_0(V))*\sum_x g(V,w) f(V,w)  \\
&- (b-a)\sum_w g(V,w) f(V,w)
\{ \frac{p(E=0|V)}{\hat{p}(E=0|V)}-1   \}
\{  p(W=w | V, E=0)- \hat{p}(W=w | V, E=0)  \} \\
&+ g(X)f(X) \{ (\mu_1(V)-\hat{\mu}_1(V))(\frac{1}{\hat{\tau}_d(X)}- \frac{1}{\tau_d(X)}) \}
    \\
&-g(X)f(X) \{ (\mu_0(V)-\hat{\mu}_0(V))(\frac{1}{\hat{\tau}_d(X)}- \frac{1}{\tau_d(X)}) \}
    \\
&- (b-a) g(X) f(X)
 (\hat{p}(W \mid V, E=0) -p(W \mid V, E=0))(\frac{1}{\hat{\tau}_d(X)}- \frac{1}{\tau_d(X)}) 
\big] \\
\end{align*}
Equivalently

\begin{align*}
    & \mathbb{E} \big[
 \sum_w  g(V,w) f(V,w)(b-a+\hat{\tau}(V,w)) 
\{ \frac{p(E=0|V)}{\hat{p}(E=0|V)}-1   \} 
\{   \hat{p}(W=w | V, E=0)-p(W=w | V, E=0)  \}
\\& +\sum_w \hat{\tau}_n g(V,w) f(V,w) \frac{1}{\hat{\tau}_d(V,w)^2 }
\big( \hat{\tau}_d(V,w)-\tau_d(V,w)
    \big)^2
    +\\&
    \frac{\pi_1-\hat{\pi}_1}{\hat{\pi}_1}(\mu_1(V)-\hat{\mu}_1(V))
    *\sum_w g(V,w) f(V,w) 
    \\
    & -  \frac{\pi_0-\hat{\pi}_0}{\hat{\pi}_0}(\mu_0(V)-\hat{\mu}_0(V))*\sum_w g(V,w) f(V,w)  \\
&+ g(X)f(X) \{ (\mu_1(V)-\hat{\mu}_1(V))(\frac{1}{\hat{\tau}_d(X)}- \frac{1}{\tau_d(X)}) \}
    \\
&-g(X)f(X) \{ (\mu_0(V)-\hat{\mu}_0(V))(\frac{1}{\hat{\tau}_d(X)}- \frac{1}{\tau_d(X)}) \}
    \\
&- (b-a) g(X) f(X)
 (\hat{p}(W \mid V, E=0) -p(W \mid V, E=0))(\frac{1}{\hat{\tau}_d(X)}- \frac{1}{\tau_d(X)}
\big] \\
\end{align*}
Bounding the Bias: The Bias is bounded by (using cauchy-shwarz)

\begin{align*}
    & \mathbb{E} \big[
 \sum_w  g(V,w) f(V,w)(b-a+\hat{\tau}(V,w) ) \\
&\| \{ \frac{p(E=0|V)}{\hat{p}(E=0|V)}-1   \} \|_2
\| \{   \hat{p}(W=w | V, E=0)-p(W=w | V, E=0)  \} \|_2
\\& +\sum_w \hat{\tau}_n g(V,w) f(V,w) \frac{1}{\hat{\tau}_d(V,w)^2 }
\| \big( \hat{\tau}_d(V,w)-\tau_d(V,w)
    \big)\|_2^2
    +\\&
    \| \frac{\pi_1-\hat{\pi}_1}{\hat{\pi}_1} \|_2 \|(\mu_1(V)-\hat{\mu}_1(V)) \|_2
    *\sum_w g(V,w) f(V,w) 
    \\
    & -  \| \frac{\pi_0-\hat{\pi}_0}{\hat{\pi}_0} \|_2 \| (\mu_0(V)-\hat{\mu}_0(V)) \|_2*\sum_w g(V,w) f(V,w)  \\
&+  g(X)f(X) \|  (\mu_1(V)-\hat{\mu}_1(V)) \|_2 \| (\frac{1}{\hat{\tau}_d(X)}- \frac{1}{\tau_d(X)})  \|_2
    \\
&-g(X)f(X) \|  (\mu_0(V)-\hat{\mu}_0(V)) \|_2 \|(\frac{1}{\hat{\tau}_d(X)}- \frac{1}{\tau_d(X)}) \|_2
 \\
&- (b-a) g(X) f(X)
\|\hat{p}(W \mid V, E=0) -p(W \mid V, E=0)\|_2 \| ({\frac{1}{\hat{\tau}_d(X)}}- \frac{1}{\tau_d(X)}) \|_2 
\big] \\
\end{align*}
Setting $\nu(V,w) =p(W=w| V, E=0) $,$\rho_0(V)= p(E=0|V)$, it becomes clear our bias is second order and bounded below by some constants
Thus our theorem where we get

    \begin{align} 
    R_n &= \mathbb{P}\{\varphi(X; \beta, \hat{\eta}) - \varphi(X; \beta, \eta_0)\} \\
&\mathrel{\substack{\le \\ \sim}} \| {\hat{\rho}}- \rho    \|_2
\|   \hat{\nu}-\nu   \|_2 +
\|  \hat{\nu}- \nu
    \|_2^2
    + \| \hat{\pi}_1 -\pi_1\|_2 \|\hat{\mu}_1 -\mu_1 \|_2  \\
     &+  \| \hat{\pi}_0-\pi_0 \|_2 \| \hat{\mu}_0 - \mu_0 \|_2
     + \|  \mu_1-\hat{\mu}_1 \|_2 \|  \hat{\nu}-\nu \|_2 
    +
  \| \hat{\mu}_0- \mu_0 \|_2 \| \hat{\nu}-\nu\|_2 
    \end{align}
\subsection{Simulation Setup Details}
We follow the following procedure to create our synthetic dataset for simulation:

We start by sampling \( n = 10,000 \) observations of 6 covariates \( \mathbf{V} \). Three of these covariates are continuous and three are discrete.

1. Covariates Generation
\begin{align*}
\mathbf{V}_{\text{continuous}} &\sim \mathcal{N}(\mu = 1, \sigma^2 = 0.5^2) \\
\mathbf{V}_{\text{discrete}} &\sim \text{Bernoulli}(p = 0.5) \\
\mathbf{V} &= [\mathbf{V}_{\text{continuous}}, \mathbf{V}_{\text{discrete}}]
\end{align*}

2. Generating Discrete Variables \textbf{\(\mathbf{W}\)}
\[
\mathbf{W}_i = \left[W_{i1}, W_{i2}, W_{i3}\right]
\]
where each \( W_{ij} \) is generated using a logistic function:
\[
W_{ij} \sim \text{Bernoulli}\left( \sigma\left(\mathbf{V}_i \cdot \mathbf{\beta}_j\right) \right)
\]
Here, \( \sigma(x) = \frac{1}{1 + e^{-x}} \) is the logistic function, and \( \mathbf{\beta}_j \) are random coefficients:
\[
\mathbf{\beta}_j \sim \text{Uniform}(-1, 1)
\]

3. Generating \( E \) and \( A \)
We generate the binary variable \( E \) based on the covariates \(\mathbf{V}\):
\[
E_i \sim \text{Bernoulli}\left( \sigma\left( \sum_{k=1}^{6} V_{ik} \cdot \gamma_{Ek} \right) \right)
\]
where \(\gamma_{Ek} \sim \text{Uniform}(-1, 1)\) are random coefficients and \(\sigma(x) = \frac{1}{1 + e^{-x}}\) is the logistic function.

Similarly, we generate the binary variable \( A \):
\[
A_i \sim \text{Bernoulli}\left( \sigma\left( \sum_{k=1}^{6} V_{ik} \cdot \gamma_{Ak} \right) \right)
\]
where \(\gamma_{Ak} \sim \text{Uniform}(-1, 1)\) are random coefficients.

4. True Conditional Treatment Effect:
To generate the true conditional treatment effect for each observation, we compute \(\beta_A\) as a linear combination of \(\mathbf{V}\) and \(\mathbf{W}\):
\[
\beta_{Ai} = \sum_{k=1}^{6} V_{ik} \cdot \alpha_{Vk} + \sum_{j=1}^{3} W_{ij} \cdot \alpha_{Wj}
\]
where \(\alpha_{Vk} \sim \text{Uniform}(0, 1.5)\) and \(\alpha_{Wj} \sim \text{Uniform}(0, 1.5)\) are random coefficients.

5. Outcome Variable \( Y \):
The outcome variable \( Y \) is computed as:
\[
Y_i = \beta_{Ai} \cdot A_i + \sum_{j=1}^{3} W_{ij} \cdot \beta_{Wj} + \sum_{k=1}^{6} V_{ik} \cdot \beta_{Vk} + \epsilon_i
\]
where \(\beta_{Wj} \sim \text{Uniform}(1, 3)\) and \(\beta_{Vk} \sim \text{Uniform}(1, 3)\) are random effect sizes, and \(\epsilon_i \sim \mathcal{N}(0, 1)\) is a random noise component. 

The simulations do not require heavy compute or anything beyond a personal machine. 1 iteration of the simulation may take 1-2 minutes to run on a personal laptop.
\subsection{Application: Licorice Gargle}
This dataset is from a study conducted by \citep{gargle} which tested the hypothesis that gargling with licorice solution immediately before induction of anesthesia prevents sore throat and postextubation coughing in patients intubated with double-lumen tubes. In this study, a total of 236 adult patients who were scheduled for elective thoracic surgery that required the use of a double-lumen endotracheal tube were recruited. The dataset includes information on various patient characteristics such as gender, physical status, body mass index (BMI), age, Mallampati score (a measure of airway visibility), smoking status, preoperative pain, and the size of the surgery. The intervention received by each patient is also recorded. The outcomes of interest, which include the presence of cough, sore throat, and pain during swallowing, were assessed at different time points. The dataset has been thoroughly cleaned and is complete, with only two patients missing outcome data. We manually remove these two patients from consideration. Therefore, finally, number of subjects $N=234$. No outliers or other data issues were identified in the dataset.

Patients were randomly allocated to one of two groups: the first group received 0.5 grams of licorice, while the second group received 5 grams of sugar, which was chosen to match the sweetness of the licorice solution. The patients' condition was assessed at multiple time points. At each assessment, the severity of sore throat was measured using an 11-point Likert scale, where 0 indicated no pain and 10 represented the worst possible pain. The presence and severity of cough were also evaluated at these time points. Additionally, pain during swallowing was assessed using the same 11-point Likert scale 30 minutes after the patients' arrival in the PACU (post-anethesia care unit). For the purpose of this study, sore throat was considered present when the patient reported a score greater than 0 on the visual analog scale.

For our experiment, we consider the following variables as $W$(present only in the target population data):
\begin{itemize}
    \item "extubation$\_$cough": This binary variable represents the presence or absence of coughing right after the removal of the endotracheal tube (extubation).
\end{itemize}
We consider the following variables as $V$(present in both study and target population data):
\begin{itemize}
    \item "preOp$\_$calcBMI": This variable represents the calculated Body Mass Index (BMI) of the patients before the surgery (preoperative).
    \item "preOp$\_$age": This variable represents the age of the patients before the surgery.
    \item "preOp$\_$pain": This variable represents the presence or severity of pain experienced by the patients before the surgery on an 11 point Likert scale from 0 to 10.
    \item "preOp$\_$mallampati": This variable represents the Mallampati score of the patients before the surgery. The Mallampati score is a measure of airway visibility and is used to predict the difficulty of intubation.
    \item "preOp$\_$asa": This variable represents the American Society of Anesthesiologists (ASA) physical status classification of the patients before the surgery. The ASA physical status is a measure of the patient's overall health and is used to assess the risk of anesthesia and surgery.
    \item "preOp$\_$smoking": This binary variable represents the smoking status of the patients before the surgery.
\end{itemize}
We use "pacu30min$\_$swallowPain", the pain in swallowing after 30 minutes, a measurement on a Likert scale from 0 to 10, as the outcome variable $Y$.

\end{document}